\documentclass[lettersize,journal]{IEEEtran}
\usepackage{amsmath,amsfonts}
\usepackage{algorithmic}
\usepackage{algorithm}
\usepackage{subfigure}
\usepackage{array}
\usepackage[caption=false,font=normalsize,labelfont=sf,textfont=sf]{subfig}
\usepackage{textcomp}
\usepackage{stfloats}
\usepackage{url}
\usepackage{verbatim}
\usepackage{graphicx}
\usepackage{cite}
\usepackage{color}
\usepackage{lineno}
\begin{document}	


\title{HGAN: Hierarchical Graph Alignment Network for Image-Text Retrieval}

\author{Jie Guo, \IEEEmembership{Member,~IEEE,} Meiting Wang, Yan Zhou,  Bin Song, \IEEEmembership{Senior Member,~IEEE,}\\ Yuhao Chi, \IEEEmembership{Member,~IEEE,} Wei Fan, Jianglong Chang
\thanks{Jie Guo, Meiting Wang, Yan Zhou, Bin Song and Yuhao Chi are with the State Key Laboratory of Integrated Services Networks, Xidian University, Xi'an, Shaanxi 710071, China. (e-mails: jguo@xidian.edu.cn, mtw@stu.xidian.edu.cn, zhouyan123y@stu.xidian.edu.cn, bsong@mail.xidian.edu.cn, yhchi@xidian.edu.cn.)}
\thanks{Wei Fan and Jianglong Chang are with the Guangdong OPPO Mobile Telecommunications Corp., Ltd, China. (e-mails: richard.fan@oppo.com, changjianglong@oppo.com.) }
\thanks{Bin Song and Yuhao Chi are both the corresponding authors.}
}
 
\markboth{Journal of \LaTeX\ Class Files,~Vol.~14, No.~8, August~2021}%
{Shell \MakeLowercase{\textit{et al.}}: A Sample Article Using IEEEtran.cls for IEEE Journals}


\maketitle

\begin{abstract}
Image-text retrieval (ITR) is a challenging task in the field of multimodal information processing due to the semantic gap between different modalities. In recent years, researchers have made great progress in exploring the accurate alignment between image and text. However, existing works mainly focus on the fine-grained alignment between image regions and sentence fragments, which ignores the guiding significance of context background information. Actually, integrating the local fine-grained information and global context background information can provide more semantic clues for retrieval. In this paper, we propose a novel Hierarchical Graph Alignment Network (HGAN) for image-text retrieval. 
First, to capture the comprehensive multimodal features, we construct the feature graphs for the image and text modality respectively. Then, a multi-granularity shared space is established with a designed Multi-granularity Feature Aggregation and Rearrangement (MFAR) module, which enhances the semantic corresponding relations between the local and global information, and obtains more accurate feature representations for the image and text modalities. Finally, the ultimate image and text features are further refined through three-level similarity functions to achieve the hierarchical alignment.
To justify the proposed model, we perform extensive experiments on MS-COCO and Flickr30K datasets. Experimental results show that the proposed HGAN outperforms the state-of-the-art methods on both datasets, which demonstrates the effectiveness and superiority of our model.
\end{abstract}


\section{Introduction}
\IEEEPARstart{I}{n} recent years, with the rapid growth of multimedia data, multimodal information processing has become more and more important. As the two most commonly-used modalities, the image and text have prompted many researchers to study cross-modal tasks, including cross-modal retrieval \cite{Qian2022AdaptiveLG}, visual question answering \cite{Liu2021VisualQA}, image captioning \cite{Yu2022DualAO}, etc. In particular, image-text retrieval (ITR) \cite{Wang2021PFANBI} task focuses on measuring the semantic similarity between images and texts.
Although great progress has been made in recent years, the heterogeneous differences caused by the inconsistent forms of images and texts seriously hinder the performance of ITR in complex scenarios, thus the ITR task still remains a great challenge.
To establish the intrinsic connection between images and texts, early works such as \cite{Frome2013DeViSEAD} extract image and text features separately using existing visual and language models used for other tasks, then directly convert them to the same dimensions as global representations. After that they measure the similarity of the global representations in this shared space, and subsequently optimize the model parameters based on matching facts.
 However, these methods only focus on global matching, that is, the alignment of the whole image and the whole sentence, and do not fully consider the potential relationship between the regions of image and the words of sentence, resulting in limited performance improvement and low interpretability. Due to the development of language representation models such as BERT \cite{Devlin2019BERTPO}, fine-grained features of text are readily available. Meanwhile, for the image modality, object detection methods such as Faster-RCNN \cite{Ren2015FasterRT} have also shown superior performance in extracting the fine-grained features. SCAN \cite{SCAN} is the first attempt to introduce object detection method into ITR task, and point out that there is an underlying alignment relationship between image regions and sentence fragments, which triggered a lot of researches on the fine-grained image-text alignment. 
For example, by generating a guidance vector from the initially extracted fine-grained features, an adaptive feature optimization is accomplished in \cite{Wehrmann2020AdaptiveCE}, which modifies the representation of the fine-grained features in another modality.Aiming to achieve more precise semantic alignment, some researchers optimize feature representations by attention mechanisms, such as \cite{SHAN} to discriminate negative pairs with similar semantic content but slightly different contextual information, improve information interactions between modalities based on cross-attention, and employ a multi-level alignment strategy with progressive matching to acquire more complementary and adequate semantic cues. Other researchers dedicate to exploring additional cues using graph convolutional networks, as in \cite{VSRN}, they construct a vision graph from region features, infer the relationship between regions using GCN, and then input node features to GRU to produce more discriminative image features.\par
\begin{figure}[tb]
	\centering
	\includegraphics[width=0.5\textwidth]{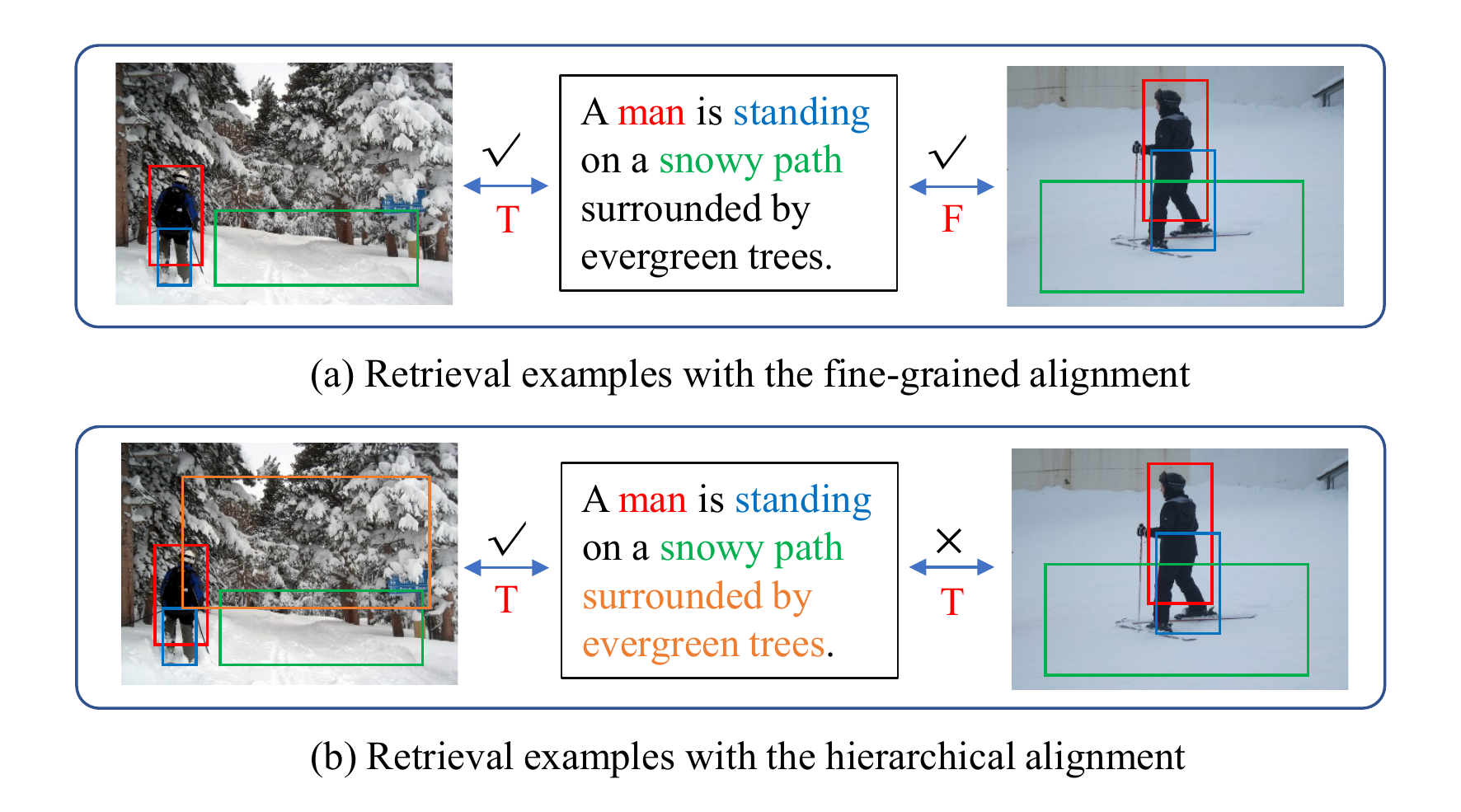}
	\caption{An illustration of retrieval examples of the fine-grained and hierarchical image-text retrieval. (a) represents the retrieval examples of fine-grained alignment, in which objects and relationships between images and texts are matched, but the result is wrong. (b) represents the retrieval examples of hierarchical alignment, which considers background context information and can distinguish negative samples with similar semantics but slightly different background. Note that ``T'' and ``F'' denote the correct and wrong matching between the image and text, respectively.}
	\label{Introduction}
\end{figure}

These methods have greatly promoted the progress of the ITR task. However, some intractable problems still exist. When studying the fine-grained alignment, many researchers ignore the importance of non-object elements such as the context background information. As shown in Fig. \ref{Introduction} (a), the fine-grained alignment method might match the sentence ``A man is standing on a snowy path surrounded by evergreen trees'' with both the left and right images since they both contain the objects ``man'', ``snow path'', and the relation ``standing''. But the matching result with the image on the right is wrong due to the difference in the context background information of the image and text.
Therefore, ignoring non-object elements such as context background information may cause false matching results, especially for negative samples with similar objects but slightly different context background. 
Accordingly, this motivates us to explore the hierarchical alignment approach shown in Fig. \ref{Introduction} (b), where hierarchical alignment refers to multi-level alignment with different granularity features, that is, not only aligning fine-grained features but also taking into account the correspondence between coarse-grained features and mixed coarse-fine-grained intermediate granularity features.
This strategy of focusing on multi-granularity characteristics has yielded excellent results in applications such as object recognition \cite{multigran}, visual classification \cite{VC}, and text-based question answering \cite{TA}.
In the task of image-text matching, this method fully considers object elements and non-object elements, which encourages more accurate matching of images and texts, and reduces the impact of the tremendous useless information existing in semantic alignment.
Regarding the intermediate granularity, one way is to fuse coarse-grained  and fine-grained features as hybrid multi-granularity features \cite{unified}, and another way is to obtain features of different granularity by setting different sizes of the visual field to the feature extraction network \cite{gran}. The first method is straightforward and direct, but it necessitates the design of a suitable fusion process, whereas the second one is affected by the choice of the size of each visual field and demanding that the feature extraction module is not a black box.
Moreover, researchers usually aggregate the features of image and text modality through max pooling \cite{Karpathy2015DeepVA} or average pooling \cite{Chen2020ExpressingOJ} in the vision-language shared space, which both ignore the importance of the synergistic relationship of local object and global context. Specifically, these methods learn the local and global features separately and do not consider the impact of the object-context-fused information on ITR task.
In summary, existing methods have two notable problems. First, most researchers consider the semantic alignment of images and texts from one perspective, such as the fine-grained alignment or global alignment, and rarely take multi-granularity feature fusion into account.
Second, the existing feature aggregation approaches in the shared semantic space, which largely ignore the object-context information interaction of multimodal features, need to be further improved.

To address these issues, this paper proposes a novel Hierarchical Graph Alignment Network (HGAN) for image-text retrieval, which establishes a multi-granularity shared space with multi-granularity feature aggregation and rearrangement (MFAR) module and performs hierarchical image-text alignment through three-level similarity functions.
Specifically, we first fuse the global and local features of the image, and construct the feature graphs for the image and text modality respectively. 
The feature graphs can preserve relative positional relationships, which is beneficial to explore the comprehensive multimodal representation.
Then, a multi-granularity shared space is established with multi-layer MFAR module, which optimizes the image and text features through feature aggregation and feature rearrangement to accomplish multi-granularity feature fusion.
The MFAR module can filter out the noisy parts of the global-local semantic alignment and retain the dominate parts to enhance the semantic alignment with object-context-fused information.
Finally, we design three similarity functions corresponding to three levels of fine-grained feature, unified feature and multi-granularity feature. The ultimate image and text features are further refined through three-level similarity functions to achieve hierarchical alignment in the multi-granularity shared space.

The main contributions of this paper are summarized as follows:

\begin{itemize}
	\item{} We establish a multi-granularity shared space with the designed Multi-granularity Feature Aggregation and Rearrangement (MFAR) module to achieve multi-granularity feature fusion. The MFAR module explores the multi-granularity feature denoising, which is dedicated to filter out the noisy parts and retain the dominate parts of the object-context-fused information.
	\item{} We propose a novel Hierarchical Graph Alignment Network (HGAN) to achieve the multi-level image-text alignment. The HGAN model aligns image and text features through multiple similarity functions to further improve the matching accuracy in the multi-granularity shared space.
	\item{} The proposed HGAN outperforms state-of-the-art image-text retrieval methods on several benchmark datasets, e.g. MS-COCO (1K and 5K) and Flickr30K.
\end{itemize}

%
%

The rest of this paper is organized as follows. In Section \uppercase\expandafter{\romannumeral2}, we introduce the related work. Then all the details of the proposed method and the experiments are presented in Section \uppercase\expandafter{\romannumeral3} and Section \uppercase\expandafter{\romannumeral4}. Finally, the conclusion is described in Section \uppercase\expandafter{\romannumeral5}.

\section{Related Work}
In this part, we will further introduce the related work including image-text retrieval and graph convolution network.
\subsection{Image-Text Retrieval}
Image-text retrieval aims at calculating the similarity between the images and texts, which can be mainly divided into three categories, global matching, regional matching and multi-level matching methods
 Global matching methods normally embed the global features of images and texts into a common embedding space, and then employ a ranking loss calculating the distance between image-text pair. Kiros et al. \cite{Kiros2014UnifyingVE} utilize the convolution neural network (CNN) for image encoding as well as the recurrent neural network (RNN) for text encoding, and Faghri et al. \cite{VSE++} employ the ResNet152 and GRU modules to encode images and texts respectively. 
Regional matching methods cast lights on the fine-grained alignment of the regions of image and the words of sentence. 
Qu at ul.\cite{CAMERA} adopt pyramid dilated convolution to obtain a multi-view representation from image region features, using the viewpoint that best matches the text feature to measure similarity.
Chen et al. \cite{GPO} design a generalized pooling operator to replace average pooling and max pooling, which can adaptively aggregate visual and language features to improve the performance of cross-modal retrieval.
Multi-level matching methods consider multi-granularity feature alignment of image and text, or matching of global feature, region feature, and other intermediate feature. The second is similar to the ontology depicted in \cite{Ontology}, whose hierarchical levels map different granularity of features.
Qi et al. \cite{Qi2018CrossmediaMA} consider not only the global and local alignments but also the relation alignment across images and texts, which can learn more precise cross-modal relevance.
Zeng et al. \cite{HSLM} propose a multi-layer graph convolutional network with object-level, object-relational-level, and higher-level learning sub-networks to learn hierarchical semantic correspondences by both local and global alignment.
Huang et al. \cite{Huang2019BiDirectionalSA} propose a bi-directional spatial-semantic attention network, which uses the word-to-regions relation to deduce the most relevant image regions, and employ the visual object to words relation to infer the close words for visual objects in the images. 
Recently, Diao et al. \cite{SGRAF} use similarity graph reasoning and similarity attention filtration module to reason about the relationship between global-local alignment information, focusing on more informative alignments and have achieved SOTA performance.
\begin{figure*}[ht]
	\centering
	\includegraphics[width=1\textwidth]{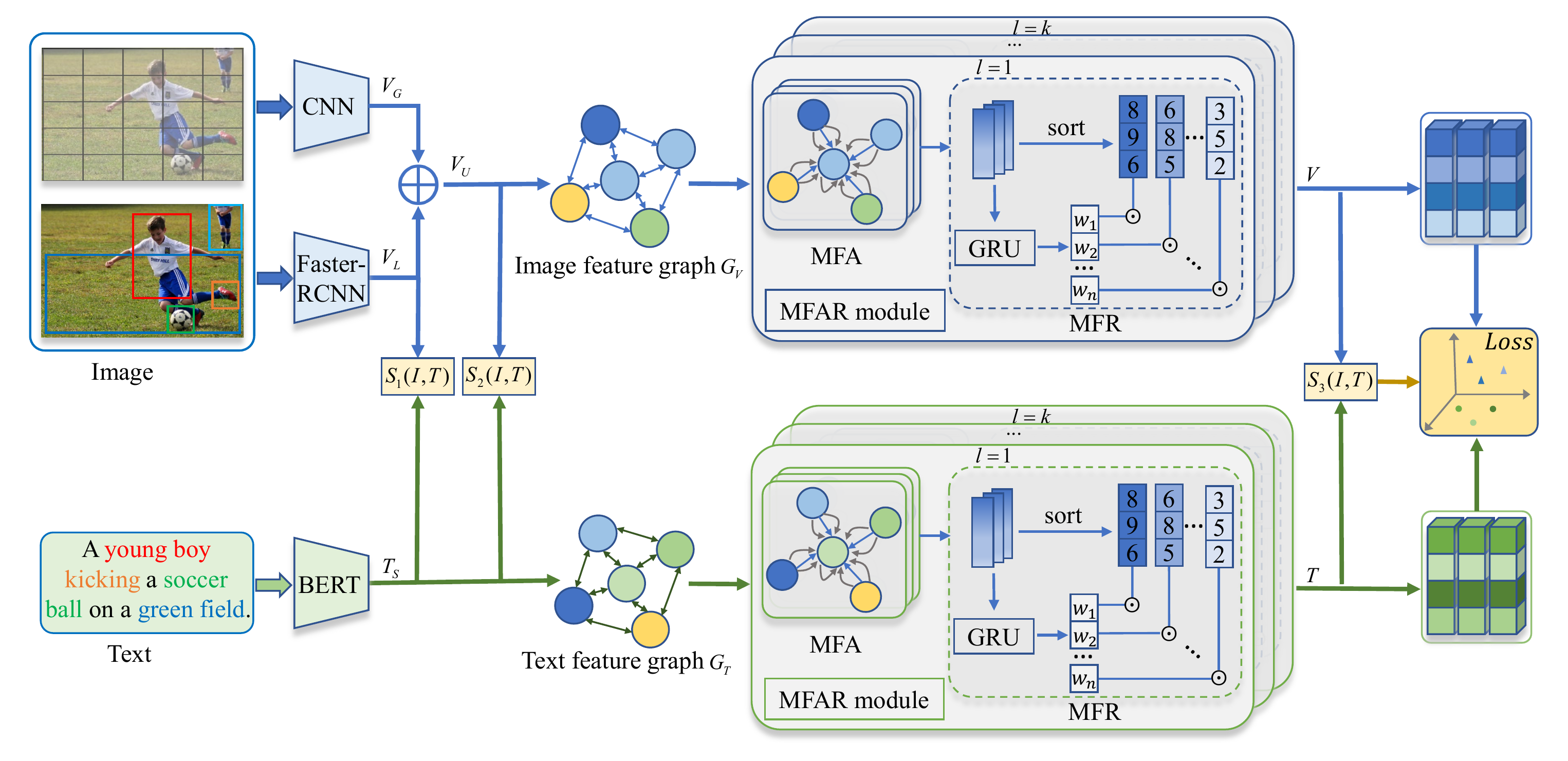}
	\caption{An illustration of the proposed HGAN model, which consists of four parts. (1) Image Feature Graph. The global and local features are extracted using ResNet152 and Faster-RCNN respectively, and then the image feature graph is constructed. (2) Text Feature Graph. The text feature is extracted using the BERT, and then the text feature graph is constructed. (3) Multi-granularity Feature Aggregation and Rearrangement Module. We design a multi-granularity feature aggregation and rearrangement module to aggregate global and local features by multi-granularity feature fusion. (4) Loss Function. The image and text features are optimized through three similarity functions $S_{1}$, $S_{2}$ and $S_{3}$.}
	\label{HGAN}
\end{figure*}

\subsection{Graph Convolution Network}
Graph convolution network (GCN) has been successfully applied into many tasks in the cross-modal fields such as cross-modal retrieval\cite{Yu2020ReasoningOT,  Zhang2022AggregationBasedGC}, image captioning \cite{Chen2020SayAY, Wang2020MultimodalAW}, visual question answering \cite{Song2018FromPT, Jiang2020InDO} and visual entailment \cite{Wang2019NeighbourhoodWR, Deng2018VisualGV}. Rather than merely focus on the similarities in content, GCN can discover the potential semantic relationships among different modalities and integrate the information of neighbor nodes. As a result, GCN is adopted into the field of cross-modal retrieval to learn the representations across different modalities, as it will be more accurate when learning local and stationary features on graphs. Yu et al. \cite{Yu2018ModelingTW} adopt GCN to enhance the representation of text features, combining the strengths of structural information with semantic information. Li et al. \cite{VSRN} utilize GCN in feature reasoning to generate image features with semantic relationship between image regions, as response of each node is computed based on its neighbour nodes. Yao et al. \cite{Yao2018ExploringVR} utilize a GCN-based encoder to refine the representations of each region proposed on objects with the learned region-level features. Yang et al. \cite{Yang2019VisualSN} delve into the details of improving semantic navigation using GCN by incorporating prior knowledge and updating it dynamically as the agent receives the current environment information. Wen et al. \cite{DSRAN} design the dual semantic relation module through the graph attention network (GAT) \cite{GAT}, which aims to enhance the regional and global relations for more accurate visual and text representations.
Liu et al. \cite{GSMN} design the graph structure for both the image and text to perform node-level and structure-level matching.

\section{Proposed Method}
In this section, we present the proposed Hierarchical Graph Alignment Network (HGAN) in detail. As shown in Fig. \ref{HGAN}, the model consists of four parts: the image feature graph (Section \uppercase\expandafter{\romannumeral3}. A),
the text feature graph (Section \uppercase\expandafter{\romannumeral3}. B), the multi-granularity feature aggregation and rearrangement module (Section \uppercase\expandafter{\romannumeral3}. C) and the loss function (Section \uppercase\expandafter{\romannumeral3}. D). 
\subsection{Image Feature Graph}
Since there exist abundant semantic information in images, we try to learn image representations by jointly considering the global and local features.
Different feature encoders are employed to obtain the global and local features respectively, and then a unified image representation is obtained by using concatenate operation to construct the image feature graph. 

\subsubsection{Global Image Representation}
We use ResNet152 \cite{ResNet152} as an encoder for extracting the global feature of images.
It is a model trained on ImageNet \cite{ImageNet} to accurately extract the pixel-level features of the image. We discard the last fully connected layer of the ResNet152 and perform the reshape operation to get the output features,
$G=\{{{g}_{1}},{{g}_{2}},...,{{g}_{m}}\},{{g}_{i}}\in {{R}^{{{D}^{0}}}}$, where ${{D}^{0}}$ represents the dimension of each pixel, and $m$ is the reshaped size of the feature map. Then we use a fully connected layer to project ${{g}_{i}}$ into the $D$-dimensional embeddings:
\begin{equation}
	{{V}_{G}}={{W}_{g}}G+{{b}_{g}},
\end{equation}
where ${{V}_{G}}$ is the global feature of the image. ${{W}_{g}}\in {{R}^{D\times {{D}^{0}}}}$ represents the weight matrix with the bias ${{b}_{g}}$, and they are learnable parameters.

\subsubsection{Local Image Representation}
In order to enable a feature vector encoding a salient region, we use the bottom-up attention \cite{Anderson2018BottomUpAT}.
Following the feature extraction method in \cite{SCAN,VSRN}, we employ the Faster-RCNN \cite{Ren2015FasterRT} as an encoder to extract the region feature of images. It is a model pre-trained on Visual Genomes DataSet \cite{VGC} to accurately identify objects in the image. 
Specifically, the features of image regions are employed to represent the image.
Therefore, the output of the image region feature encoder can be represented as $L=\{{{l}_{1}},{{l}_{2}},...,{{l}_{k}}\},{{l}_{i}}\in {{R}^{{{D}^{0}}}}$, where ${{D}^{0}}$ represents dimension of region feature, and $k$ is the number of detected regions. For each selected region $i$, we use the average pooling layer to get ${{l}_{i}}$. Then we use a fully connected layer to project ${{l}_{i}}$ into the $D$-dimensional embeddings:
\begin{equation}
	{{V}_{L}}={{W}_{l}}L+{{b}_{l}},
\end{equation}
where ${{V}_{L}}$ is the local feature of the image. ${{W}_{l}}\in {{R}^{D\times {{D}^{0}}}}$ represents the weight matrix, and ${{b}_{l}}$ represents the bias.

\subsubsection{Unified Image Representation}
Previous works separately considered on the global and local features of the images, and the important semantic relationship between the local object and global context is ignored.
Differently, we design a unified representation of the global and local features of the images, where coarse-grained and fine-grained information is fused. And the unified image representation is constructed in the form of graph, so as to use graph convolutional networks to optimize multi-granularity feature fusion, thus enhance it to learn more accurate image features.
By using concatenate operation, ${{V}_{G}}$ and ${{V}_{L}}$ form a unified image representation ${{V}_{U}}$:
\begin{equation}
	{{{V}_{U}}={{V}_{G}}\|{{V}_{L}}},
\end{equation}
where $\|$ denotes the concatenate operation. So we can get the region-pixel unified feature ${{V}_{U}}=\{v_{U}^{1},v_{U}^{2},...,v_{U}^{m+k}\}, v_{U}^{i}\in {{R}^{{{D}^{U}}}}$. Then the relationship  ${{E}_{U}}$  between $v_{U}^{i}$ and $v_{U}^{j}$ is defined:
\begin{equation}
	{{E}_{U}}(v_{U}^{i},v_{U}^{j})=v_{U}^{i}\odot v_{U}^{j},
\end{equation}
where $\odot $ represents the element product. Finally, the features $v_{U}^{i}$ and $v_{U}^{j}$ are regarded as the nodes, and the relationship ${{E}_{U}}$ between them is regarded as the edge to construct the image feature graph ${{G}_{V}}=({{V}_{U}},{{E}_{U}})$.

\subsection{Text Feature Graph}

The traditional text feature representation methods use the RNN-based models such as LSTM \cite{LSTM} or GRU \cite{GRU}, and treat the output of the network as the sentence feature.
The language representation model BERT \cite{Devlin2019BERTPO} uses the self-attention based transformer structure, which is accomplished at learning semantic relationships.
The model has powerful feature extraction capabilities to generate deep bidirectional linguistic representations for word tokens.
In this paper, we employ the BERT as the text encoder. First, the sentence is tokenized by WordPiece, and then the features of the word are extracted through the BERT model.
Therefore, the output of the text encoder can be represented as $S=\{{{s}_{1}},{{s}_{2}},...,{{s}_{l}}\},{{s}_{i}}\in {{R}^{{{D}^{1}}}}$, where ${{D}^{1}}$ represents the dimension, and $l$ is the maximum number of words in the sentence.
Then we use a fully connection layer to project ${{t}_{i}}$ into the $D$-dimensional embeddings:
\begin{equation}
	{{T}_{S}}={{W}_{s}}S+{{b}_{s}},
\end{equation}
where ${{T}_{S}}$ is the feature of the texts. ${{W}_{s}}\in {{R}^{D\times {{D}^{1}}}}$ represents the weight matrix, and ${{b}_{s}}$ represents the bias. The text feature can be represented as ${{T}_{S}}=\{t_{S}^{1},t_{S}^{2},...,t_{S}^{l}\},t_{S}^{i}\in {{R}^{{{D}^{S}}}}$. Then in order to build the text feature graph, we define the relationship ${E}_{S}$ between $t_{S}^{i}$ and $t_{S}^{j}$:
\begin{equation}
	{{E}_{S}}(t_{S}^{i},t_{S}^{j})=t_{S}^{i}\odot t_{S}^{j},
\end{equation}
where $\odot $ represents the element product. Finally, the features $t_{S}^{i}$ and $t_{S}^{j}$ are regarded as the nodes, and the relationship ${{E}_{S}}$ between them is regarded as the edge to construct the text feature graph ${{G}_{T}}=({{T}_{S}},{{E}_{S}})$.

\subsection{Multi-granularity Feature Aggregation and Rearrangement Module}
In this section, we introduce the designed multi-granularity feature aggregation and rearrangement (MFAR) module, which filters out the noisy parts of the global-local semantic alignment and retains the useful parts to accomplish multi-granularity feature fusion.
For image and text feature graphs, we both use the MFAR module to conduct feature aggregation and rearrangement. Here the image modality is taken as an example to illustrate the details of MFAR module.
\subsubsection{Multi-granularity Feature Aggregation Module}
The node feature representation is optimized in this module. Considering that the nodes may contain redundant information, we employ the attention mechanism to learn the correlation between nodes and selectively aggregate neighboring nodes based on the correlation to obtain optimized node features, which filter worthless information and keep effective information. Therefore, semantic interaction between global and local information is realized throughout this process.\par
Given the image feature graph ${{G}_{V}}=({{V}_{U}},{{E}_{U}})$, ${{V}_{U}}=\{v_{U}^{1},v_{U}^{2},...,v_{U}^{m+k}\},v_{U}^{i}\in {{R}^{{{D}^{U}}}}$, where $v_{U}^{i}$ is the node feature, and ${{E}_{U}}$ is the relationship of nodes. ${{D}^{U}}$ represents the dimension of image feature.
To obtain sufficient expressive power, we parameterize node features through the weight matrix, and then, use the self-attention mechanism to calculate the attention coefficient for each node:
\begin{equation}
	{{e}_{ij}}=\frac{{{W}_{q}}v_{U}^{i}\odot {{W}_{k}}v_{U}^{j}}{\sqrt{{{D}^{U}}}},
\end{equation}
where ${{e}_{ij}}$ indicates the importance of node $j$ to node $i$, ${{W}_{q}}$ and ${{W}_{k}}$ are learnable weight matrices. $\odot$ denotes the element product. 
To make the weight coefficients are comparable between different nodes, 
all choices of node $j$ are regularized using the softmax function:
\begin{equation}
	\alpha_{i j}=\operatorname{softmax}{ }_{j}\left(e_{i j}\right)=\frac{\exp \left(e_{ij}\right)}{\sum_{k \in \mathcal{N}_{i}} \exp \left(e_{i k}\right)},
\end{equation}
where ${{\mathcal{N}}_{i}}$ is the set of neighbor nodes of node $i$. At the same time, in order to reduce the number of parameters, we introduce the multi-head attention mechanism to calculate the attention coefficients, which is faster and more space-saving. 
\begin{equation}
	MultiHead\left( v_{U}^{i},v_{U}^{j} \right)={{W}_{o}}\|_{h=1}^{H}(hea{{d}_{1}},...,hea{{d}_{h}})
\end{equation}

\begin{equation}
	{{\operatorname{head}}_{h}}=\operatorname{Softmax}\left( \frac{W_{q}^{h}v_{U}^{i}\odot W_{k}^{h}v_{U}^{i}}{\sqrt{d}} \right)W_{v}^{h}v_{U}^{j},
\end{equation}
where $\|$ represents the concatenate operation, $H$ stands for the number of parallel attention layers, and $d=\frac{D}{H}$. ${{W}_{o}}\in {{R}^{D\times D}}$, ${{W}_{q}}\in {{R}^{D\times d}}$, ${{W}_{k}}\in {{R}^{D\times d}}$ and ${{W}_{v}}\in {{R}^{D\times d}}$ are learnable weight matrices. Then, the output of the multi-granularity feature aggregation module is obtained by the nonlinear activation function:
\begin{equation}
	v_{U}^{i'}=BN\left( \operatorname{ReLU}\left( \sum\limits_{j\in {{N}_{i}}}{ MultiHead}\left( v_{U}^{i},v_{U}^{j} \right) \right) \right),
\end{equation}
where $BN$ represents the batch normalization which can be used to speed up training, $ReLU$ is the activation function, and ${{\mathcal{N}}_{i}}$ is the set of neighbor nodes of node $i$. 
For the convenience of description, we simplify the output as $v^{i'}$.

\subsubsection{Multi-granularity Feature Rearrangement Module}
This module is used to better fuse multi-granularity semantics and improve multi-granularity feature representation. Given that each channel component of the node feature can be used as a representation of the node at multiple saliencies, in this module, firstly, the feature vector of the node is fine-tuned by rearranging the channel values to align the feature components of different saliencies in each node. Then, adaptive weights are learned for each node to fuse the component under varied saliency, thus optimized multi-granularity semantic features are obtained.
For the output features $v^{i'}$ of the multi-granularity feature aggregation module, most of them use average pooling to aggregate features, ignoring the information interaction between global and local features. The average pooling method obtains ${v}$ by averaging the $N$ feature vectors as ${v}=\frac{1}{N}\textstyle\sum_{i=1}^{N}{v^{i}}'$. Therefore the value at channel k is calculated by:
\begin{equation}
	{v_k}=\frac{1}{N}\sum\limits_{i=1}^{N}{v_{k}^{i'}},
\end{equation}
where $k=1,...,K$ is the channel number of ${v}$. 

We design the multi-granularity feature rearrangement to refine the feature vectors. First, we sort the feature vectors and then learn a rearrangement coefficient for each vector, taking the weighted sum of the vectors as the output:
\begin{equation}\label{vn}
    v_k=\frac{1}{N}\sum\limits_{i=1}^{N}{\theta_i}{max_k}({v^i}'),\forall k.
\end{equation}

\begin{equation}
	{{\theta }_{i}}=f[i,N],i=1,...,N.
\end{equation}
$f$ is the rearrangement coefficient generator. ${{\theta }_{i}}$ represents the rearrangement coefficient of the $i$-th node, which satisfies $\textstyle\sum_{i=1}^{\text{N}}{{{\theta }_{i}}}=1$. Therefore, the final image feature of the model can be expressed as $V=\{{{v}_{1}},{{v}_{2}},...,{{v}_{k}}\},{{v}_{i}}\in {{R}^{{{D}^{V}}}}$, where ${D}^{V}$ represents the dimension of the image feature.

Specifically, the multi-granularity feature rearrangement module consists of two components, a trigonometric function-based position encoder for generating the position indices and a bidirectional gated recurrent unit (BiGRU) based sequence model for generating rearrangement coefficients. To make full use of the prior information contained in the position indices, following \cite{Attention}, the trigonometric positional encoding strategy is employed to vectorize the positional indices:

\begin{equation}
	{{p}_{i}}(i,2j)=\sin \left( \frac{i}{{{10000}^{2j/{{d}_{p}}}}} \right),
\end{equation}

\begin{equation}
	{{p}_{i}}(i,2j+1)=\cos \left( \frac{i}{{{10000}^{2j/{{d}_{p}}}}} \right),
\end{equation}
where ${{p}_{i}}$ is the position vector, and ${{d}_{p}}$ is the dimension of the position vector. After converting the position indices into vector representations, the sequence model is adopted to generate the rearrangement coefficients:

\begin{equation}
	\left\{ {{\mathbf{\theta }}_{i}} \right\}_{i=1}^{\text{N}}=MLP\left( \operatorname{BiGRU}\left( \left\{ {{p}_{i}} \right\}_{i=1}^{\text{N}} \right) \right).
\end{equation}
Then the aggregated representation is obtained by the weighted sum, as shown in Eq.(13).
Finally, we can obtain the multi-granularity image and text representations through the multi-layer MFAR module.


\begin{table*}[ht]\scriptsize
  \centering
  \caption{The introduction of the baseline models, including the year of publication, advantages, disadvantages, and main content.}
  \setlength{\tabcolsep}{0.1cm}
  {\begin{tabular}{|m{6.3em}<{\centering}|m{2.5em}<{\centering}|m{15.25em}|m{15.25em}|m{29em}|}
    \hline
     model & year & \multicolumn{1}{m{15.25em}<{\centering}|}{pros}  & \multicolumn{1}{m{15.25em}<{\centering}|}{cons} & \multicolumn{1}{m{29em}<{\centering}|}{main content} \\
    \hline
    VSE++\cite{VSE++} & 2017  & Introduce the hard negatives  into the loss function. & Ignore local alignment of images and text. & This method introduces the hard negatives into the common loss functions used for image-text retrieval. The improved loss can better guide the more powerful image  and text encoders. \\
    \hline
    SCAN\cite{SCAN}  & 2018  &  Introduce object detection method into ITR task. & Ignore interactions between image and text modals. & This is the first method to introduce object detection method into ITR task, and point out that there is an underlying alignment relationship between image regions and sentence fragments. \\
    \hline
    CAMP\cite{CAMP}  & 2019  &  Consider fine-grained image-text interactions and adaptively control the  cross modal information flow. & Ignore abstract objects, such as descriptions of certain behaviors. & This method considers comprehensive and fine-grained image-text interactions, and handles negative pairs and unrelated information with an adaptive gating module. \\
    \hline
    RDAN\cite{RDAN}  & 2019  &  Introduce the multi-level image-text alignments into cross-modal retrieval. & Ignore the effects of intra-modal relationships. & This is a relation-wise dual attention model, which encodes the local and the global correlations between regions and words by training the image-text retrieval network. \\
    \hline
    VSRN\cite{VSRN}  & 2019  & Focus on relational reasoning in images firstly. & Only consider reasoning within the image modality, ignoring text modalities. & This method uses the gate and memory mechanism to perform global semantic reasoning on the relationship representation and gradually generate the image feature. \\
    \hline
    MMCA\cite{MMCA}  & 2020  & Explore intra-modality and the inter-modality relationship through a cross-attention mechanism. & Lack of research on semantic correspondence and semantic association. & This method designs a novel cross-attention mechanism, which exploit the intra-modality and the inter-modality relationship to enhance and complement each other for image-text retrieval. \\
    \hline
    CAAN\cite{CAAN}  & 2020  & Propose a context-aware attention network to selectively focuses on critical local fragments. & Ignore high-level semantic information between modalities. & This method selectively focuses on pivotal local features by aggregating the global context to discover latent semantic relations. \\
    \hline
    IMRAN\cite{IMRAM} & 2020  & Explore fine-grained correspondences using the attention mechanism & Ignore the alignment of phrases and image regions. & This is a fine-grained matching method, which introduces an iterative matching strategy with recurrent attention memory to explore the fine-grained alignment progressively. \\
    \hline
    GSMN\cite{GSMN}  & 2020  & Derive fine-grained image-text associations through node-level matching & Extra work to build the visual graph and textual graph. & This is a graph matching method, which models object, relation and attribute as a structured phrase to learn correspondence of object, relation,  attribute and structured phrase separately. \\
    \hline
    CAMERA\cite{CAMERA} & 2020  & Design a context-aware  multi-view summarization network to meet the  multi-view description challenge  & {Summarize text information from multiple views, easily mixed with noise.} & {This is a state-of-the-art image-text retrieval method on the Flickr30K dataset in 2020, which summarizes context-enhanced image region information from multiple views.} \\
    \hline
    Meta-SPN\cite{BFAN} & 2021  &  propose a meta self-paced network to accelerate model training. & {Just a training acceleration algorithm based on existing models.} & {This method designs a meta self-paced network, which automatically learns the weight coefficients from data for image-text retrieval.} \\
    \hline
    SMFEA\cite{SMFEA} & 2021  & Build a tree of images and texts to obtain the structured semantic representation. & {The constructed tree is coarse-grained and cannot distinguish data with high similarity.} & {This is a structured tree based image-text retrieval model, which models the relations of the image and text fragments by constructing structured tree encoders.} \\
    \hline
    SHAN\cite{SHAN} & 2021  & Realize image-text matching through multi-step cross-modal inference. & {Ignore the shared semantic concepts that potentially correlated the different modalities.} & {This is a hierarchical alignment model, which decomposes image-text retrieval into multi-step cross-modal reasoning processes.} \\
    \hline
    SGRAF\cite{SGRAF} & 2021  & Making full use of alignment information through graph inference to infer more accurate match scores. & {Change the similarity from value to vector, the retrieval process will take more time.} & {This method first applies the vector-based similarity representations to characterize the local and global features, which relies on the GCN to infer the relation-aware similarity.} \\
    \hline
    CGAM\cite{CGAM} & 2021  & consider the shared semantic concepts to enhance the discriminative power of the common space. & {Ignore syntactical alignment and other research on multi-granularity.} & {This method builds semantic-embedded graph for each modality, and smooths the discrepancy through cross-graph attention module to obtain shared semantic-enhanced features.} \\
    \hline
    CSCC\cite{CSCC}  & 2021  & Considering the syntactical correspondence through the cross-level consistency for Image-text matching. & {Ignore the effect of global context information on retrieval results.} & {This is a state-of-the-art image-text retrieval method on the MS-COCO 1K and 5K datasets in 2021, which introduces a conceptual-level image-text alignment scheme to exploring the fine-grained correspondence.} \\
    \hline
    \end{tabular}}%
  \label{tab:model}%
\end{table*}%

\subsection{Loss Function}
To align the global and local information simultaneously, we set three-level cosine similarity functions in the proposed model.
${{S}_{1}}$ calculates the similarity of image region features $V_{L}$ and text features $T_{S}$ and ${{S}_{2}}$ computes the similarity of image unified features $V_{U}$ and text features $T_{S}$. The similarity of image and text features $V$ and $T$ refined by the MFAR module are calculated by ${{S}_{3}}$. $S$ is the final similarity of image and text in the multi-granularity shared space, that is, the sum of the above three similarity functions:
\begin{equation}
	S(I,T)={{S}_{1}}({{V}_{L}},{{T}_{S}})+{{S}_{2}}({{V}_{U}},{{T}_{S}})+{{S}_{3}}(V,T),
\end{equation}
where $(I,T)$ is the matched positive pair of image and text. And the formulations for the three cosine similarity functions are specified as follows:
\begin{equation}
	{S_{1}({{V}_{L}},{{T}_{S}}) = \frac{V_{L}\cdot T_{S}}{||V_{L}||\times||T_{S}||}},
\end{equation}
\begin{equation}
	{S_{2}({{V}_{U}},{{T}_{S}}) = \frac{V_{U}\cdot T_{S}}{||V_{U}||\times||T_{S}||}},
\end{equation}
\begin{equation}
	S_{3}(V,T) = \frac{V\cdot T}{||V||\times||T||}.
\end{equation}
Then, a bidirectional hinge-based triplet ranking loss \cite{SCAN, VSE++} is adopted to make the matched image-text pairs have higher similarity scores than unmatched ones.
\begin{equation}
\begin{split}
	L={{\left[ d+S\left( I',T \right)-S(I,T) \right]}_{+}}\\+{{\left[ d+S\left( I,T' \right)-S(I,T) \right]}_{+}},
 \end{split}
\end{equation}
where $d$ denotes the margin parameter, and ${{\left[ x \right]}_{+}}\equiv \max (x,0)$. $I'=\arg {{\max }_{X\ne I}}S(X,T)$ and $T'=\arg {{\max }_{Y\ne T}}S(I,Y)$ denote the hardest negatives corresponding to the positive pair $(I,T)$.

\section{Experiment}
In this section, we evaluate the proposed HGAN model on two benchmark datasets. First, the dataset and evaluation metrics and the implementation details are introduced. Then, the effectiveness of our model is proved by the performance comparison experiments and the ablation studies. Finally, the proposed HGAN model is qualitatively analyzed through visualization experiments.
\subsection{Datasets and Evaluation Metric}
We employ two commonly-used image-text retrieval datasets, MS-COCO (1K and 5K) \cite{Lin2014MicrosoftCC} and Flickr30K \cite{Plummer2015Flickr30kEC}, to evaluate our model.

The MS-COCO is a large-scale benchmark  dataset used for image recognition, segmentation and retrieval. It is composed of 123,287 images, each with 5 corresponding captions. Following the experiment settings in \cite{SCAN, VSRN}, we evaluate our method on 1K and 5K test images respectively. Specifically, the train, validation and test splits contain 113,287, 5000 and 5000 images. The 1K branch refers to the results are reported by averaging from 5 folds of 1K test images and the 5K branch refers to testing on the full 5K test images directly.


The Flickr30K dataset contains 31,783 images. Each image is paired with 5 corresponding captions. Following the split method in \cite{VSE++} about the Flickr30K dataset, we evaluate the performance of our model using 29,000 images for training, 1,000 images for validation, and the remaining 1,000 ones for testing. 

For both image-to-text retrieval and text-to-image retrieval tasks, we report the results with the standard metrics, including R@K (Recall@K, K=1, 5, 10) and Rsum. R@K is defined as the proportion of correct image or text being retrieved among top K results, and Rsum is the sum of six R@K value to evaluate performance comprehensively.

\begin{table*}[htb]
	\centering
	\caption{Comparison with the baseline models of Image-to-Text retrieval and Text-to-Image retrieval on the 1K and 5K test set of MS-COCO dataset. The bold indicates the optimal results, and the underline indicates the suboptimal results. ``-'' denotes the results are not provided. }

		\begin{tabular}{c|ccc|ccc|c|ccc|ccc|c}
			\hline
			& \multicolumn{7}{c|}{MS-COCO 1K}                       & \multicolumn{7}{c}{MS-COCO 5K} \\
			\cline{2-15}   \multicolumn{1}{c|}{Methods} & \multicolumn{3}{c|}{Image-to-Text} & \multicolumn{3}{c|}{Text-to-Image} &       & \multicolumn{3}{c|}{Image-to-Text} & \multicolumn{3}{c|}{Text-to-Image} &  \\
			\cline{2-15}          & R@1   & R@5   & R@10  & R@1   & R@5   & R@10  & Rsum  & R@1   & R@5   & R@10  & R@1   & R@5   & R@10  & Rsum \\
			\hline
			VSE++  \cite{VSE++} & 64.6  & 90.0  & 95.7  & 52.0  & 84.3  & 92.0  & 478.6 & 41.3  & 71.1  & 81.2  & 30.3  & 59.4  & 72.4  & 355.7 \\
			SCAN  \cite{SCAN} & 72.7  & 94.8  & 98.4  & 58.8  & 88.4  & 94.8  & 507.9 & 50.4  & 82.2  & 90.0  & 38.6  & 69.3  & 80.4  & 410.9 \\
			CAMP  \cite{CAMP} & 72.3  & 94.8  & 98.3  & 58.5  & 87.9  & 95.0  & 506.8 & 50.1  & 82.1  & 89.7  & 39.0  & 68.9  & 80.2  & 410.0 \\
			RDAN  \cite{RDAN} & 74.6  & 96.2  & 98.7  & 61.6  & 89.2  & 94.7  & 515.0  & -     & -     & -     & -     & -     & -     & - \\
			VSRN  \cite{VSRN} & 76.2  & 94.8  & 98.2  & 62.8  & 89.7  & 95.1  & 516.8 & 53.0  & 81.1  & 89.4  & 40.5  & 70.6  & 81.1  & 415.7 \\
			MMCA  \cite{MMCA} & 74.8  & 95.6  & 97.7  & 61.6  & 89.8  & 95.2  & 514.7 & 54.0  & 82.5  & 90.7  & 38.7  & 69.7  & 80.8  & 416.4 \\
			CAAN  \cite{CAAN} & 75.5  & 95.4  & 98.5  & 61.3  & 89.7  & 95.2  & 515.6 & 52.5  & 83.3  & 90.9  & 41.2  & 70.3  & 82.9  & 421.1 \\
			IMRAM  \cite{IMRAM} & 76.7  & 95.6  & 98.5  & 61.7  & 89.1  & 95.0  & 516.6 & 53.7  & 83.2  & 91.0  & 39.7  & 69.1  & 79.8  & 416.5 \\
			GSMN  \cite{GSMN} & 78.4  & 96.4  & 98.6  & 63.3  & 90.1  & 95.7  & 522.5 & -     & -     & -     & -     & -     & -     & - \\
			CAMERA  \cite{CAMERA} & 77.5  & 96.3  & \underline{98.8}  & 63.4  & 90.9  & 95.8  & 522.7 & 55.1  & 82.9  & 91.2  & 40.5  & 71.7  & 82.5  & 423.9  \\
			Meta-SPN  \cite{BFAN} & 74.4  & 95.0  & 98.3  & 58.6  & 87.6  & 94.3  & 508.2 & 51.0     & 81.1     & 89.4     & 37.5     & 66.7     & 77.5     & 403.2 \\
			SMFEA  \cite{SMFEA} & 75.1  & 95.4  & 98.3  & 62.5  & 90.1  & 96.2  & 517.6 & 54.2  & -     & 89.9  & \underline{41.9}  & -     & 83.7  & 425.3 \\
			SHAN  \cite{SHAN} & 76.8  & 96.3  & 98.7  & 62.6  & 89.6  & 95.8  & 519.8 & -     & -     & -     & -     & -     & -     & - \\
			SGRAF  \cite{SGRAF} & \underline{79.6}  & 96.2  & 98.5  & 63.2  & 90.7  & 96.1  & 524.3 & \underline{57.8}  & -     & \underline{91.6}  & \underline{41.9}  & -     & 81.3  & - \\
			CGAM  \cite{CGAM} & 78.9  & \textbf{97.5}  & \underline{98.8}  & 65.7  & 90.2  & \textbf{96.6} & 527.7 & -     & -     & -     & -     & -     & -     & - \\
			CSCC  \cite{CSCC} & 78.8  & 96.1  & \textbf{99.0}  & \underline{66.6}  & \textbf{92.5}  & \underline{96.4}  & \underline{529.4}  & 55.6  & \underline{83.6}  & 91.2  & 40.8  & \underline{73.2}  & \underline{84.3}  & \underline{428.7} \\
			\hline
			\textbf{HGAN} (ours)  & \textbf{81.1} & \underline{96.9} & \textbf{99.0} & \textbf{67.4} & \underline{92.2} & \textbf{96.6} & \textbf{533.2} & \textbf{60.0} & \textbf{85.8} & \textbf{92.8} & \textbf{45.4} & \textbf{75.3} & \textbf{85.1} & \textbf{444.4} \\
			\hline
	\end{tabular}%
	\label{tab:result}%
\end{table*}%
\subsection{Comparable Methods}
In order to prove the effectiveness of the proposed HGAN method, we choose the models shown in Table \ref{tab:model} as the baseline models.  In the Table \ref{tab:model}, we introduce the main works in the field of the image-text retrieval, including the publication year, the pros and cons, and the main content of each baseline model.

\begin{table}[tbp]
	\centering
	\caption{Comparison with the baseline models of Image-to-Text retrieval and Text-to-Image retrieval on the Flickr30K dataset. 
	}
	\setlength{\tabcolsep}{0.15cm}{
		\begin{tabular}{c|ccc|ccc|c}
			\hline
			&\multicolumn{7}{c}{Flickr30K} \\
			\cline{2-8}    \multicolumn{1}{c|}{Methods} & \multicolumn{3}{c|}{Image-to-Text} & \multicolumn{3}{c|}{Text-to-Image} &  \\
			\cline{2-8}          & \multicolumn{1}{c}{R@1} & \multicolumn{1}{c}{R@5} & \multicolumn{1}{c|}{R@10} & \multicolumn{1}{c}{R@1} & \multicolumn{1}{c}{R@5} & \multicolumn{1}{c|}{R@10} & \multicolumn{1}{c}{Rsum} \\
			\hline
			VSE++ \cite{VSE++} & 52.9  & 80.5  & 87.2  & 39.6  & 70.1  & 79.5  & 407.9 \\
			SCAN \cite{SCAN} & 67.4  & 90.3  & 95.8  & 48.6  & 77.7  & 85.2  & 465.0  \\
			CAMP \cite{CAMP} & 68.1  & 89.7  & 95.2  & 51.5  & 77.1  & 85.3  & 466.9 \\
			RDAN \cite{RDAN} & 68.1  & 91.0  & 95.9  & 54.1  & 80.9  & 87.2  & 477.2 \\
			VSRN \cite{VSRN} & 71.3  & 90.6  & 96.0  & 54.7  & 81.8  & 88.2  & 482.6 \\
			MMCA \cite{MMCA} & 74.2  & 92.8  & 96.4  & 54.8  & 81.4  & 87.8  & 487.4 \\
			CAAN \cite{CAAN} & 70.1  & 91.6  & 97.2  & 52.8  & 79.0  & 87.9  & 478.6 \\
			IMRAM \cite{IMRAM} & 74.1  & 93.0  & 96.6  & 53.9  & 79.4  & 87.2  & 484.2  \\
			GSMN \cite{GSMN} & 76.4  & 94.3  & 97.3  & 57.4  & 82.3  & 89.0  & 496.8 \\
			CAMERA \cite{CAMERA} & 78.0  & \underline{95.1}  & \underline{97.9}  & 60.3  & 85.9  & \underline{91.7}  & \underline{508.9}  \\
			Meta-SPN \cite{BFAN} & 72.5  & 93.2  & 96.7  & 53.3  & 80.2  & 87.2  & 483.1 \\
			SMFEA \cite{SMFEA} & 73.7  & 92.5  & 96.1  & 54.7  & 82.1  & 88.4  & 487.5 \\
			SHAN \cite{SHAN} & 74.6  & 93.5  & 96.9  & 55.3  & 81.3  & 88.4  & 490.0  \\
			SGRAF \cite{SGRAF} & 77.8  & 94.1  & 97.4  & 58.5  & 83.0  & 88.8  & 499.6  \\
			CGAM \cite{CGAM} & \underline{78.7}  & 94.5  & \underline{97.9}  & 58.2  & 83.6  & 89.6  & 502.5 \\
			CSCC \cite{CSCC} & 72.7  & 93.4  & 96.5  & \underline{61.2}  & \underline{86.7}  & 91.5  & 502.0 \\
			\hline
			\textbf{HGAN}(ours) & \textbf{80.3} & \textbf{96.5} & \textbf{98.3} & \textbf{62.3} & \textbf{87.8} & \textbf{93.1} & \textbf{518.3} \\
			\hline
	\end{tabular}}%
	\label{tab:result2}%
\end{table}%

\subsection{Implementation Details}
In this section, we describe the software and hardware configuration of the experiments in detail. Our model is evaluated in pytorch-1.7.1 with python wrapper and a machine with Intel Xeon Gold 6226R CPU, 64GB RAM, 1T SSD and NVIDIA Tesla A100 GPU. 
Specifically, for dataset Flickr30K and MS-COCO, the model is trained for 12 and 20 epochs with the adaptive moment estimation optimizer (Adam) \cite{Adam}, respectively.
The batchsize is set to 256 and 320 for Flickr30K and MS-COCO datasets, respectively.
The warmup is set to 0.1 and the learning rate is set to 0.0002 with a decay rate of 0.1 every 6 epochs. For images, the feature dimension ${{D}^{0}}$ is set to 2048 for both global and local features. The basic version of the pre-trained BERT \cite{Devlin2019BERTPO} is used to extract text features, which includes 12 layers, 12 heads, 768 hidden units, and 110M parameters, to get the text embeddings with ${{D}^{1}=768}$. The image-text shared space dimension $D$ is set to 1024.
In our model, we set the number of MFAR layers to 2 and 4 for the Flickr30K and MS-COCO datasets, respectively.

\subsection{Performance Comparison}
In this section, we show the experimental results on the MS-COCO (1K and 5K) and Flickr30K datasets in Table \ref{tab:result} and Table \ref{tab:result2}. For the sake of fairness, we directly cite the simulation results of the baseline models in their corresponding original papers. 
The bold indicates the optimal results, and the underline indicates the suboptimal results. 
Overall, our model achieves the state-of-the-art  retrieval results on the Flickr30K and MS-COCO 1K and 5K datasets.

Table \ref{tab:result} shows the performance of each model on the MS-COCO dataset. CSCC \cite{CSCC} has the optimal performance on the 1K test set, probably because it considers the syntactic alignment in addition to the fine-grained alignment, which 
is similar to our model. SGRAF \cite{SGRAF} has the best performance on the 5K test set, which utilizes a vector-based similarity representation method to deduce more accurate matching score of images and texts through the fine-grained alignment. Overall, our HGAN model shows the best performance according to the Rsum metric. As for the 1K test set of MS-COCO, our model reaches 81.1\% R@1 score and 67.4\% R@1 score on image-to-text retrieval and text-to-image retrieval respectively, both outperforming other state-of-the-art methods.
For the 5K test set, image-to-text retrieval and text-to-image retrieval achieve the best R@1 of 60.0\% and 45.4\% respectively, with 2.2\% and 3.5\% improvement over the SGRAF model. The performance of R@5 and R@10 can also be seen from the Table \ref{tab:result}, both of which achieve the optimal and suboptimal results.

Table \ref{tab:result2} shows the performance of each model on the Flickr30K dataset. CAMERA \cite{CAMERA} has the best performance on image-to-text retrieval, which aggregates context-enhanced visual information from multiple views of the image. 
CSCC \cite{CSCC} has the best performance on the text-to-image retrieval, which  simultaneously considers the semantic information of the concept and syntactic. It is obvious that our HGAN model outperforms existing models by a large margin, achieving 80.3\%, 96.5\% and 98.3\% for R@1, R@5 and R@10 on image-to-text retrieval. And the performance on text-to-image retrieval is 62.3\%, 87.8\% and 93.1\% for R@1, R@5 and R@10 respectively. Compared to the CAMERA method, our model has a significant improvement in image-to-text retrieval and text-to-image retrieval tasks (by   2.3\% and 2\% on R@1).
In summary, our HGAN model outperforms other state-of-the-art models on the Flickr30K dataset.

\begin{table}[tbp]
	\centering
	\caption{Effectiveness analysis of the MFAR module and unified fatures on the MS-COCO 1K and Flickr30K datasets.  ``GIE'' denotes the global image embedding and ``LIE'' denotes the local image embedding.  ``MFA'' is the multi-granularity feature aggregation module.``MFAR'' is the multi-granularity feature aggregation  and rearrangement module.}
	\begin{tabular}{c|ccc|ccc}
		\hline
		& \multicolumn{6}{c}{MS-COCO 1K} \\
		\cline{2-7}    Methods & \multicolumn{3}{c|}{Image-to-Text} & \multicolumn{3}{c}{Text-to-Image} \\
		\cline{2-7}          & R@1   & R@5   & R@10  & R@1   & R@5   & R@10 \\
		\hline
		MFA+GIE & 73.4  & 94.5  & 97.9  & 60.4  & 89.7  & 95.6 \\
		MFAR+GIE & 75.4  & 94.9  & 98.3  & 62.8  & 89.9  & 95.4 \\
		MFA+LIE & 76.5  & 95.4  & 98.4  & 63.2  & 90.0    & 95.1 \\
		MFAR+LIE & 80.0    & 96.6  & 98.9  & 66.8  & 91.8  & \textbf{96.6} \\
		MFA+GIE+LIE & 78.3  & 95.7  & 98.6  & 64.5  & 90.9  & 95.9 \\
		\hline
		MFAR+GIE+LIE & \textbf{81.1} & \textbf{96.9} & \textbf{99.0} & \textbf{67.4} & \textbf{92.2} & \textbf{96.6} \\
		\hline
		& \multicolumn{6}{c}{Flickr30K} \\
		\hline
		MFA+GIE & 75.4  & 92.6  & 96.3  & 57.4  & 84.1  & 90.7 \\
		MFAR+GIE & 76.1  & 92.7  & 96.9  & 57.4  & 84.9  & 90.9 \\
		MFA+LIE & 77.8  & 94.4  & 97.5  & 60.8  & 87.0    & 92.6 \\
		MFAR+LIE & 79.5  & 95.2  & 97.9  & 61.9  & 87.4  & 92.8 \\
		MFA+GIE+LIE & 78.6  & 94.2  & 97.3  & 61.4  & 87.2  & 92.8 \\
		\hline
		MFAR+GIE+LIE & \textbf{80.3} & \textbf{96.5} & \textbf{98.3} & \textbf{62.3} & \textbf{87.8} & \textbf{93.1} \\
		\hline
	\end{tabular}%
	\label{tab:addlabel3}%
\end{table}%

\begin{table}[tbp]
	\centering
	\caption{Effectiveness analysis of the hierarchy on the MS-COCO 1K and Flickr30K datasets.}
	\begin{tabular}{c|ccc|ccc}
		\hline
		& \multicolumn{6}{c}{MS-COCO 1K} \\
		\cline{2-7}    \multicolumn{1}{c|}{Methods} & \multicolumn{3}{c|}{Image-to-Text} & \multicolumn{3}{c}{Text-to-Image} \\
		\cline{2-7}          & R@1   & R@5   & R@10  & R@1   & R@5   & R@10 \\
		\hline
		${{S}_{3}}$ & 78.8  & 96.2  & 98.8  & 65.1  & 92.0  & \textbf{96.6}\\
		${{S}_{1}}+{{S}_{3}}$ & 79.8  & 96.5  & 98.9  & 66.4  & 91.9  & 96.5 \\
		${{S}_{2}}+{{S}_{3}}$ & 79.7  & 96.5  & 98.8  & 66.2  & \textbf{92.5} & 96.4 \\
		\hline
		${{S}_{1}}+{{S}_{2}}+{{S}_{3}}$ & \textbf{81.1} & \textbf{96.9} & \textbf{99.0} & \textbf{67.4} & 92.2  & \textbf{96.6} \\
		\hline
		& \multicolumn{6}{c}{Flickr30K} \\
		\hline
		${{S}_{3}}$ & 79  & 95.4  & 97.6  & 61.1  & 87.0  & 92.7 \\
		${{S}_{1}}+{{S}_{3}}$ & 79.9  & \textbf{96.6}  & \textbf{98.5}    & 62.0    & \textbf{87.8}  & 93.0 \\
		${{S}_{2}}+{{S}_{3}}$ & 79.5  & 96.1  & 98.1  & 61.4  & 87.1  & 92.4 \\
		\hline
		${{S}_{1}}+{{S}_{2}}+{{S}_{3}}$ & \textbf{80.3} & 96.5 & 98.3 & \textbf{62.3} & \textbf{87.8} & \textbf{93.1} \\
		\hline
	\end{tabular}%
	\label{tab:addlabel4}%
\end{table}%

\subsection{Analysis of Model}
In this section, the ablation experiments are performed on the MS-COCO 1K and Flickr30K datasets. We analyze the effectiveness of the MFAR module and the unified features, the influence of the different similarity functions and the parameters of the model, respectively.

\emph{1) Effect of the MFAR module.}
In Table \ref{tab:addlabel3}, the MFAR module is the proposed multi-granularity feature aggregation and rearrangement. The MFA module denotes the multi-granularity feature aggregation module and the MFR module is the multi-granularity feature rearrangement module. To demonstrate the effectiveness of the MFAR module, we disable the MFR module for performance testing to explore the impact of the designed modules. We can observe that our MFAR module has always outperformed the MFA-based model, which brings about a 2\% performance promotion on two benchmark datasets. 

\emph{2) Effect of the unified features.}
In Table \ref{tab:addlabel3}, GIE denotes the global image embedding and LIE denotes the local image embedding. GIE+LIE is the unified image representation.
We can observe that global matching using GIE module is not as effective as fine-grained matching using LIE module, and further, using the unified image feature including GIE and LIE module in our HGAN model has the optimal performance. Specifically, comparing the results of `MFAR+GIE' and `MFAR+LIE', it can be found that using only local features to construct image features is much better than using only global features, indicating that local features describe more image details and can generate more discriminative features. 
Furthermore, the comparison of the results of `MFAR+LIE' and `MFAR+GIE+LIE' shows that better results can be achieved by using both global and local features, which means that the global features supplement the contextual information lacked by the local features, allowing the reconstructed features to better represent the integral image and achieve improved retrieval results.

\emph{3) Effect of hierarchy.}
The hierarchy in our proposed method is directly reflected by multiple levels of similarity. Specifically, the three similarity functions ${{S}_{1}}$, ${{S}_{2}}$, and ${{S}_{3}}$ corresponding to three levels of fine-grained feature, unified feature and multi-granularity feature, respectively.
In Table \ref{tab:addlabel4}, ${{S}_{1}}$ represents the similarity function between the image local feature and text feature, and ${{S}_{2}}$ works on the image unified feature and text feature. ${{S}_{3}}$ cannot be removed in our HGAN model, which represents the similarity function between image and text after optimization by MFAR module.  The bold font indicates the optimal results. We can find that the application of multi-level similarity function achieves about 1.5\% improvement on R@1 for image-to-text retrieval and text-to-image retrieval.
In terms of performance improvement, ${{S}_{1}}$ is more effective compared to ${{S}_{2}}$, since ${{S}_{1}}$ takes into account the alignment of local features.

\begin{figure}[ht]
	\centering
	\subfigure[MS-COCO 1K.]
	{\includegraphics[width=0.24\textwidth]{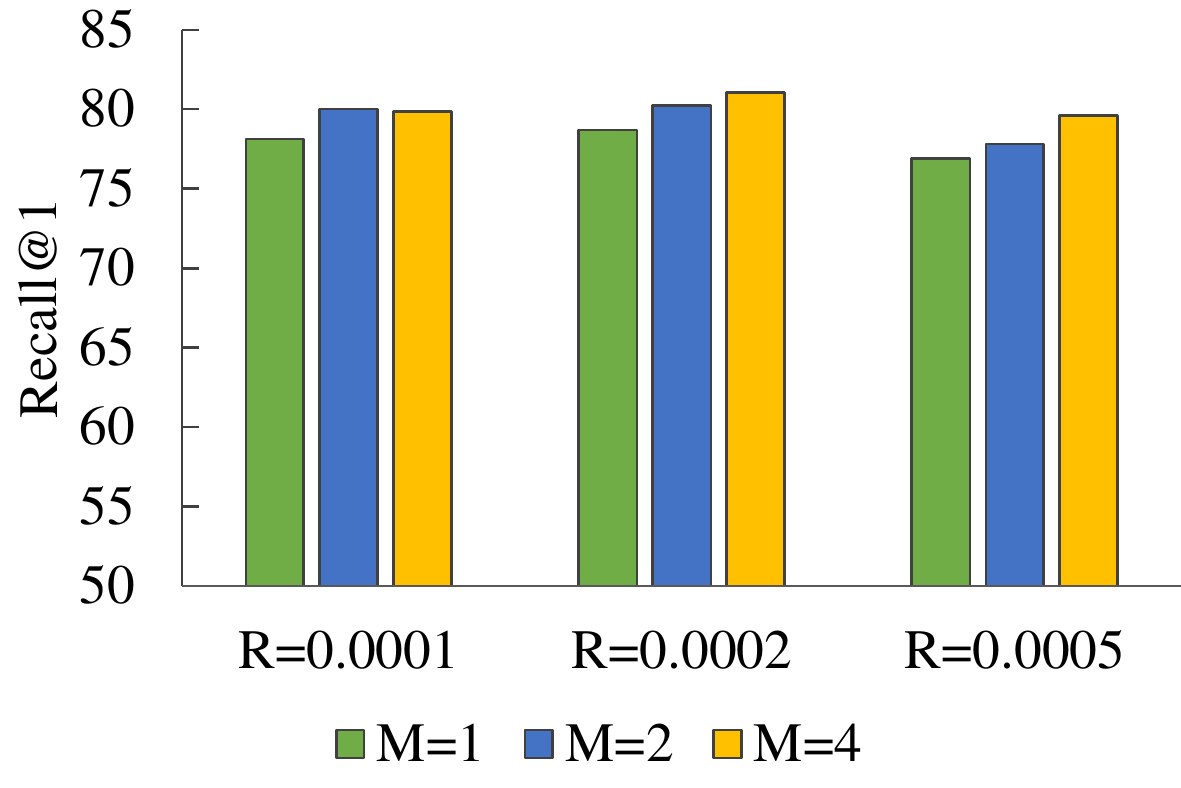}}
	\subfigure[Flickr30K.]
	{\includegraphics[width=0.24\textwidth]{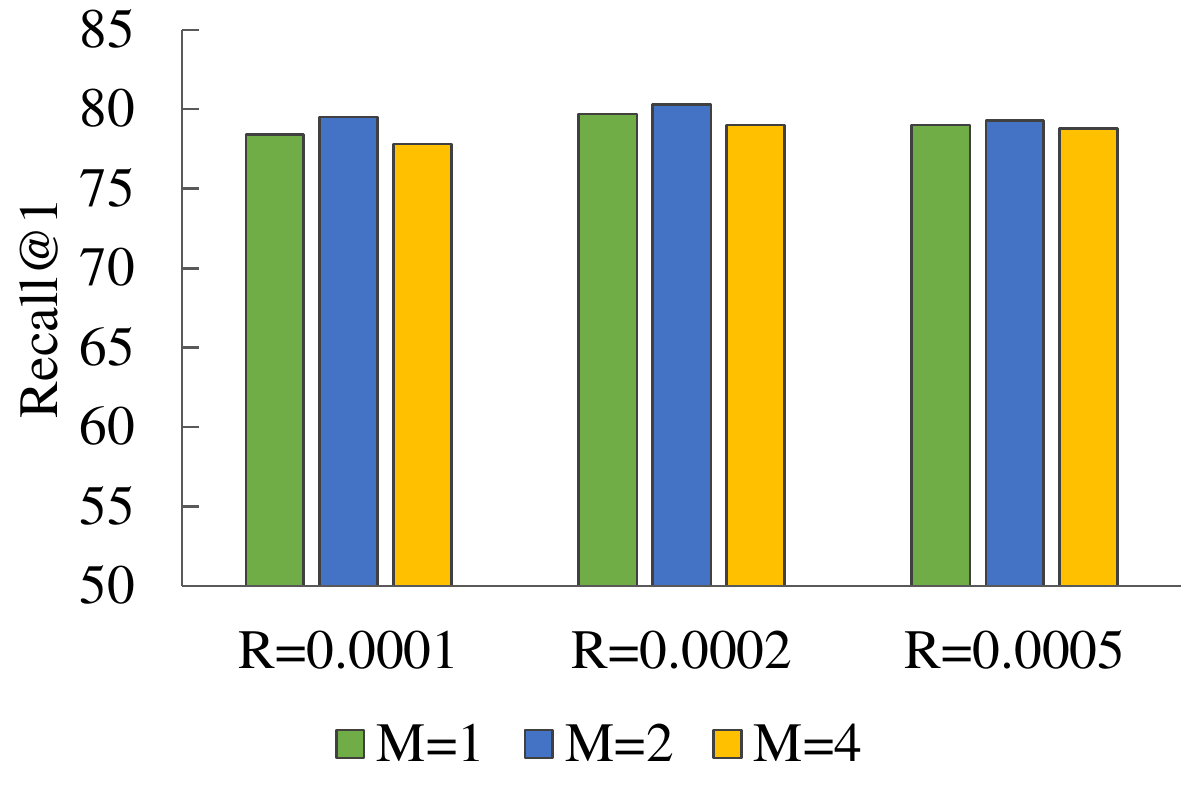}}
	 \caption{The results of Recall@1 with different initial learning rate and MFAR module layers.}
	\label{M}
\end{figure}

\begin{figure}[ht]
	\centering
	\includegraphics[width=0.35\textwidth]{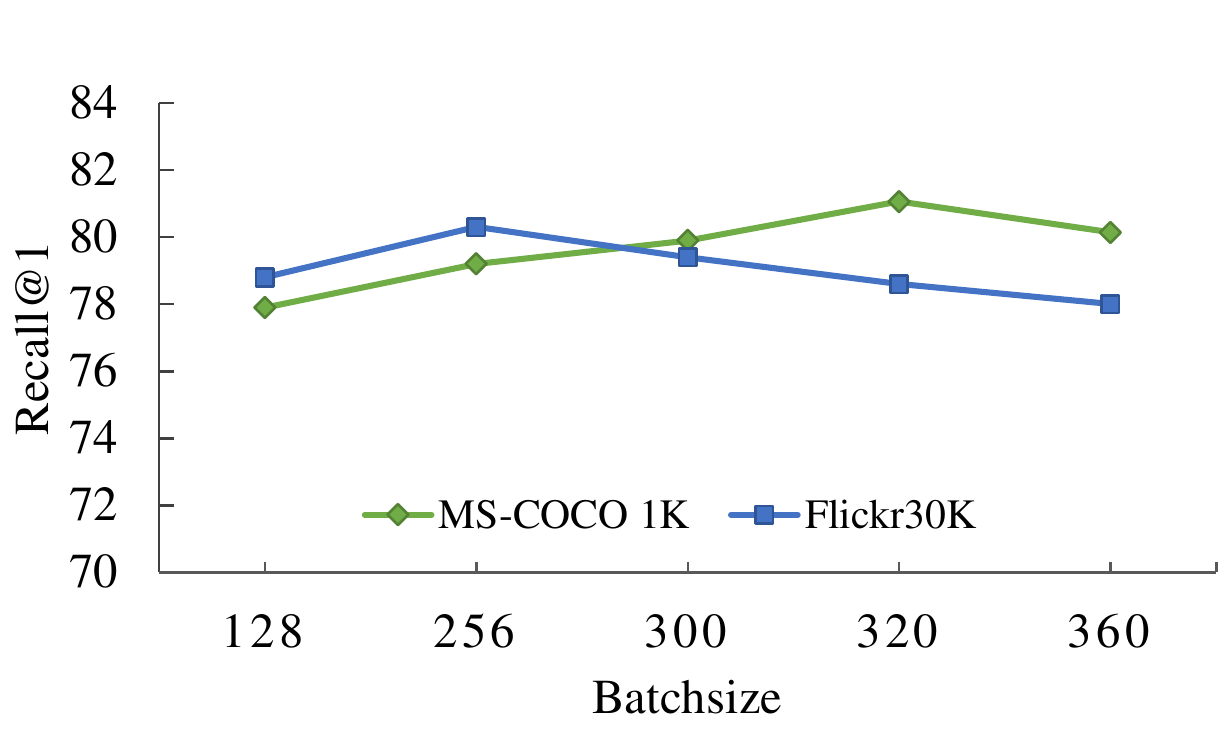}
	 \caption{The results of Recall@1 with different batchsize.}
	\label{batch}
\end{figure}

\emph{4) Effect of the parameters.}
In this section, we analyze the impact of model parameters on performance, including the number of layers of the MFAR module, the learning rate, and the batchsize. In Fig. \ref{M}, $M$ represents the number of layers of the MFAR module, and R denotes the initial learning rate. First, we can see that for the MS-COCO dataset, the performance is improved as the number of layers of the MFAR module increases, and it has the best performance when $M=4$. Due to equipment limitations, we have no way to conduct experiments with $M=8$, but we think that $M=4$ has achieved the ideal performance. For the Flickr30K dataset, $M=2$ works best because the two-layer MFAR module is sufficient to extract the information contained in the dataset. Also, setting an initial learning rate of 0.0002 is the most appropriate for several datasets of our model. Furthermore, Fig. \ref{batch} shows the optimal batchsize on several benchmark datasets. The best batchsize value is 320 for the MS-COCO dataset, and for the Flickr30K dataset, the optimal batchsize is 256.

\begin{figure}[htbp]
	\centering
	\subfigure[MS-COCO 1K.]
	{\includegraphics[width=0.24\textwidth]{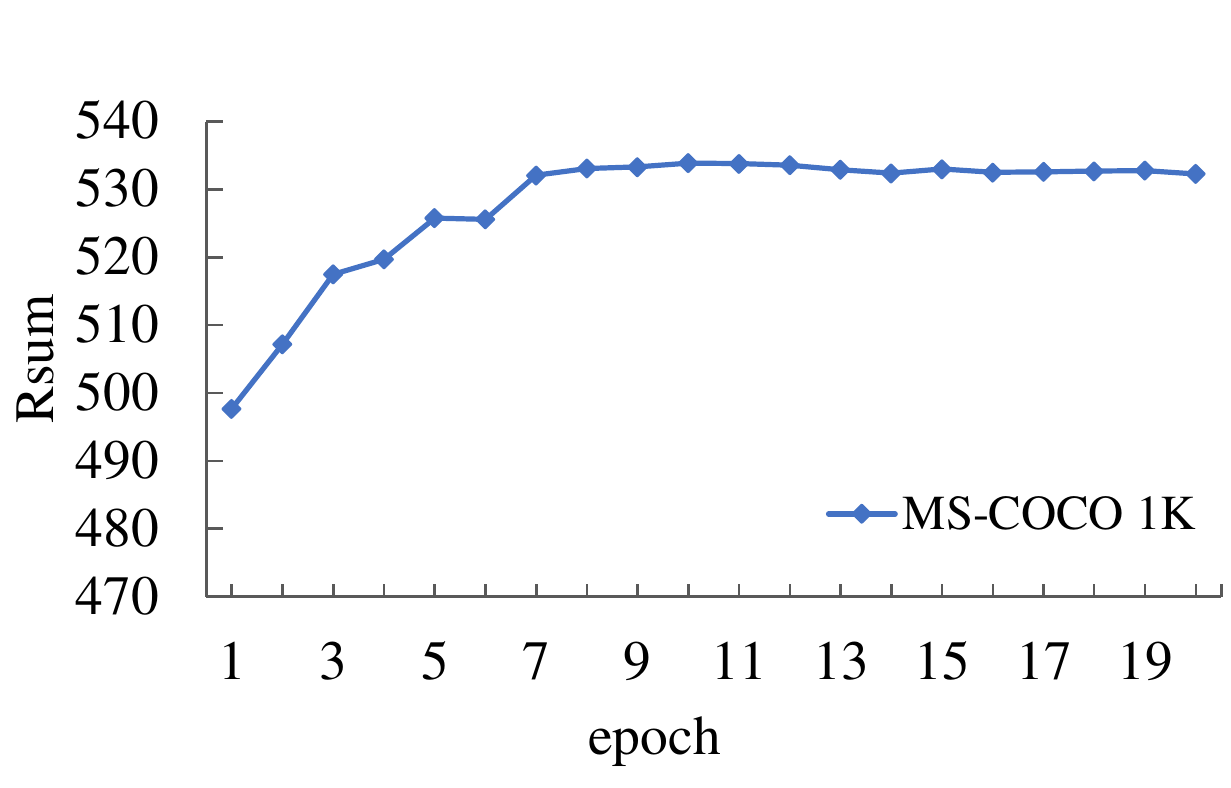}}
	\subfigure[Flickr30K.]
	{\includegraphics[width=0.24\textwidth]{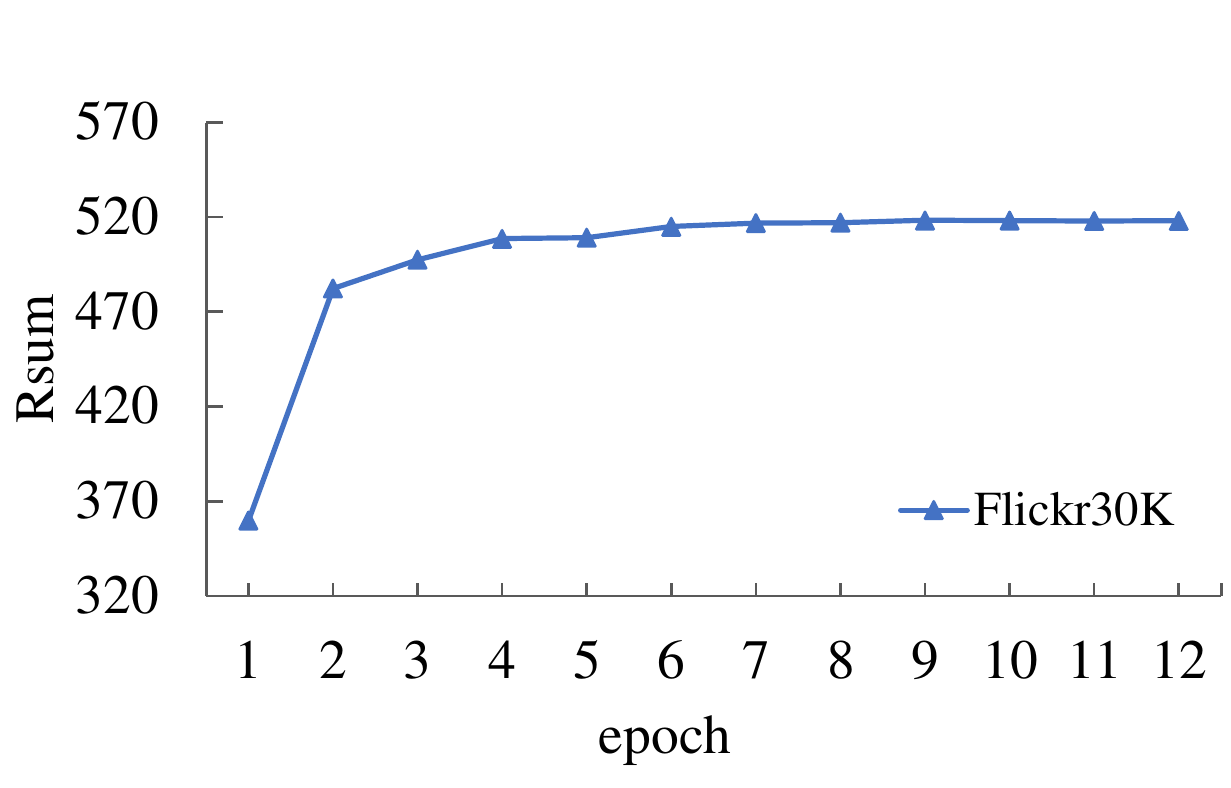}}
	 \caption{The values of Rsum when varying the number of epochs.}
	\label{epoch}
\end{figure}

\begin{figure}[htbp]
	\centering
	\subfigure[MS-COCO 1K.]
	{\includegraphics[width=0.24\textwidth]{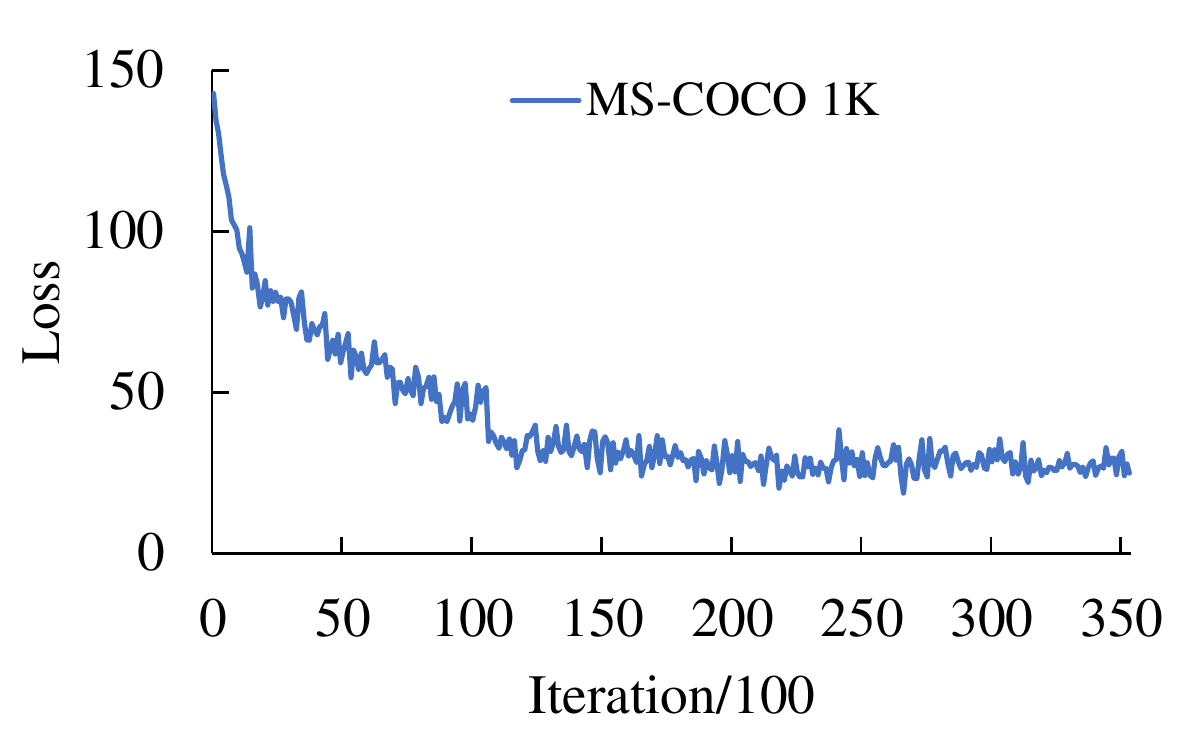}}
	\subfigure[Flickr30K.]
	{\includegraphics[width=0.24\textwidth]{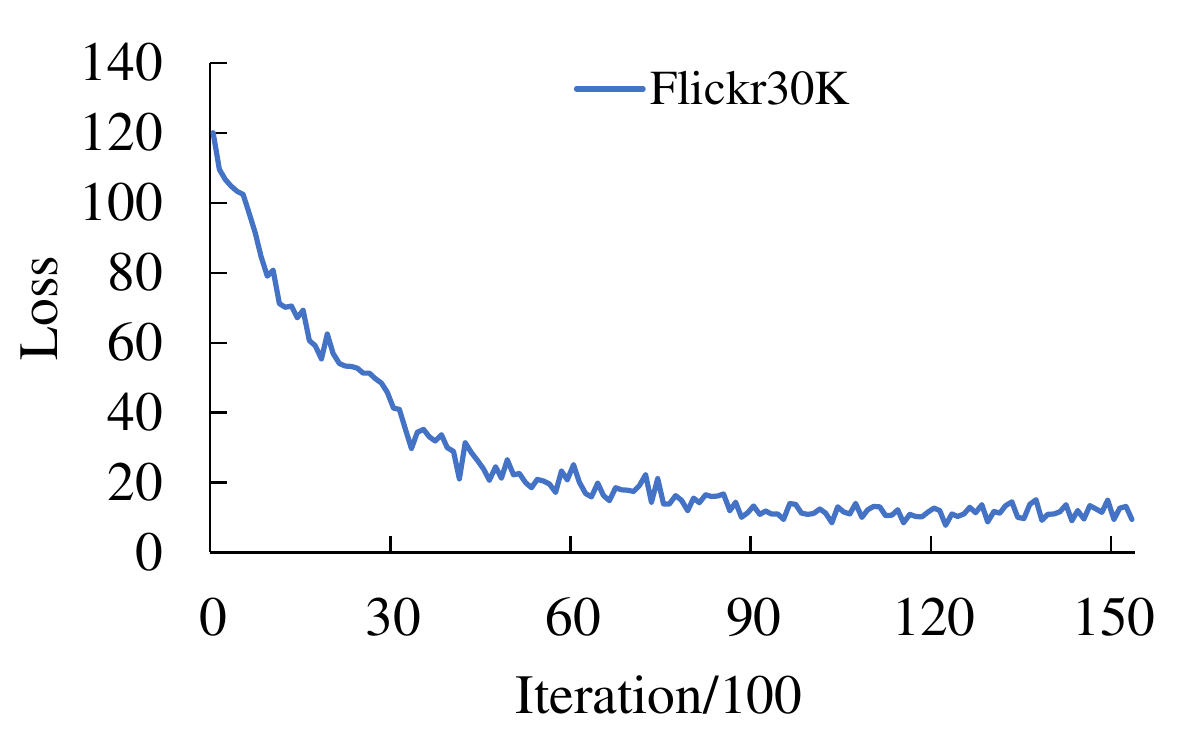}}
	\caption{The variation of loss values with different iteration numbers.}
	\label{Iteration}
\end{figure}

\subsection{Analysis of Training Process}
In this section, we analyze the training process of the proposed HGAN model. Firstly, in order to measure the computational efficiency of our model, we consider the computational complexity of the model (measured by FLOPs) and the time complexity (measured by the number of parameters), which are 36.81G and 211.29M respectively. Compared to the trillions of parameters of large-scale pre-trained models in the image-text cross-modal domain, our approach achieves reasonable calculation consumption. \par
Subsequently, we analyze the change of Rsum values (the sum of recall value) with epoch and the change of loss values with iteration. 
Fig. \ref{epoch} records the values of Rsum when varying the number of epochs on the MS-COCO 1K and Flickr30K test sets. The maximum value of the epoch is
set to 20 and 12 for the MS-COCO 1K and Flickr30k datasets, respectively. For MS-COCO 1K, when the epoch reaches about 10, Rsum reaches the maximum and then stabilizes. For Flickr30k, the model has basically converged when epoch is 6 because it contains less image and text.
Fig. \ref{Iteration} records the variation of loss values with different iteration numbers on the MS-COCO 1K and Flickr30K test sets. 
It can be seen that our model can converge to a satisfactory value on both the MS-COCO 1K and Flickr30K, and there is no underfitting and overfitting.
Comparing the datasets MS-COCO 1K and Flickr30K, MS-COCO 1K needs more epochs of training because of its large amount of data and complex information. Nonetheless, the proposed model has better performance in MS-COCO 1k dataset because of the richer relations contained in.

\begin{figure*}[htbp]
	\centering
	\includegraphics[width=1\textwidth]{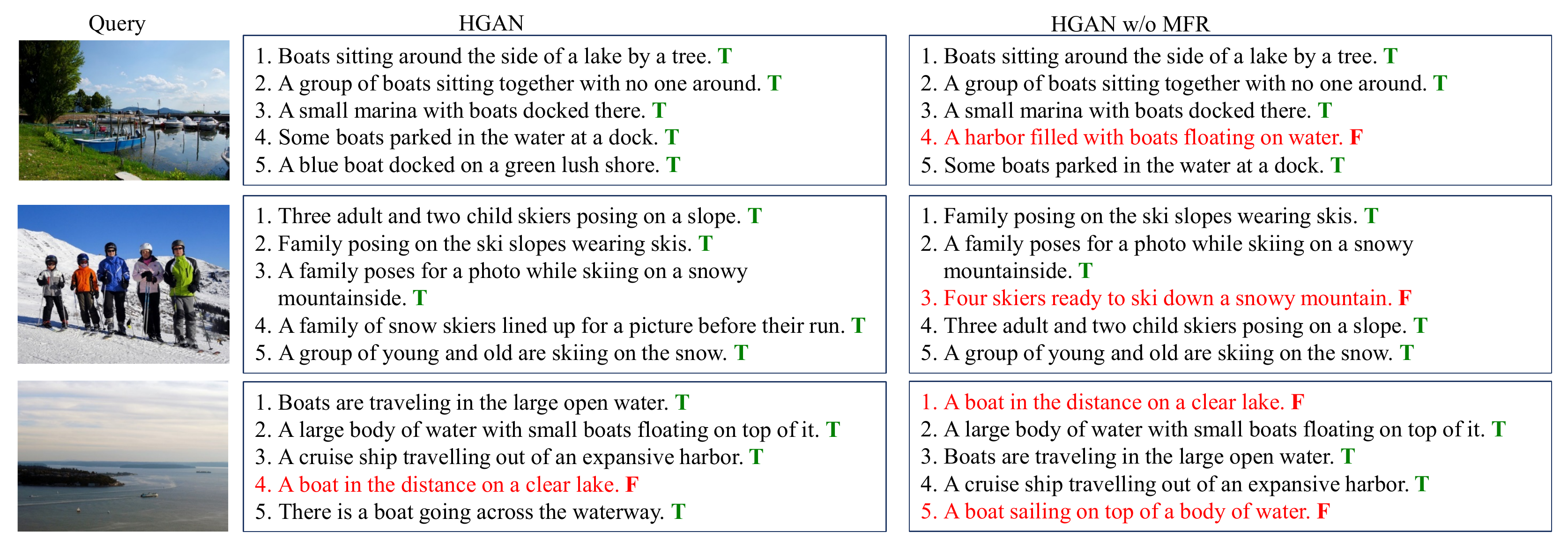}
	\caption{The qualitative results of image-to-text retrieval on the MS-COCO 1K dataset. The top-5 retrieval results are shown for each query image. The green ``T'' denotes the correct sentences and the red ``F'' indicates the wrong sentences (best viewed in color).}
	\label{figure3}
\end{figure*}

\begin{figure*}[htbp]
	\centering
	\includegraphics[width=1\textwidth]{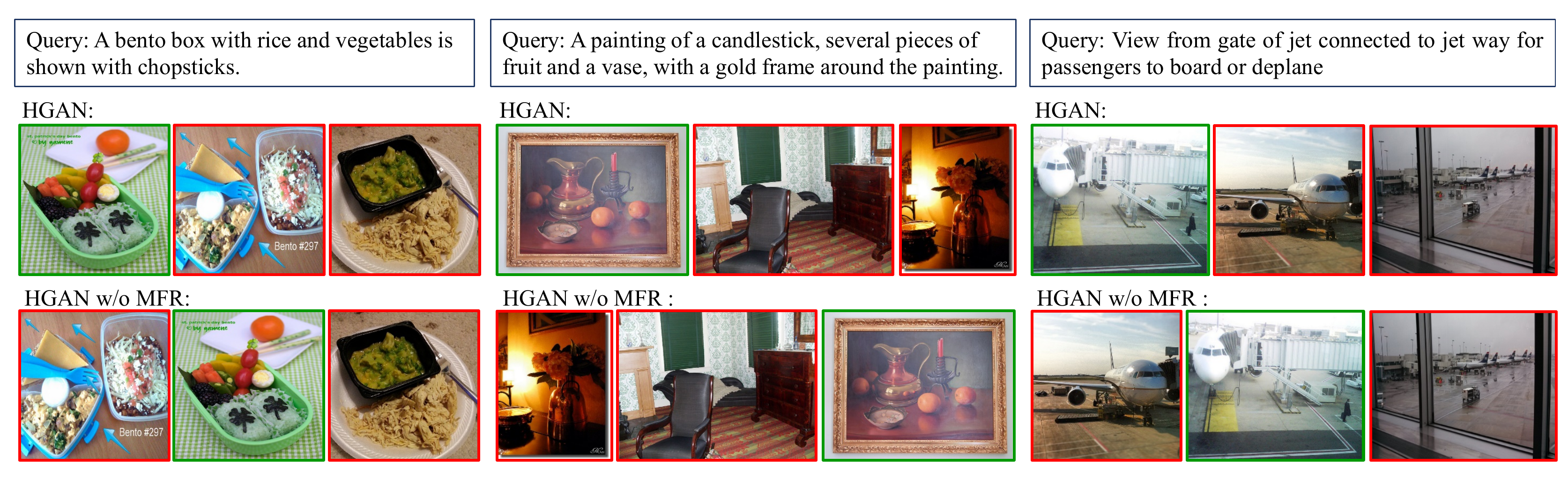}
	\caption{The qualitative results of text-to-image retrieval on the MS-COCO 1K dataset. The top-3 retrieval results are shown for each query text. The correct images are highlighted in green boxes, and the wrong images are highlighted in red boxes (best viewed in color).}
	\label{figure4}
\end{figure*}

\subsection{Visualization of Retrieval Results}
In this section, we discuss the qualitative results of the proposed HGAN model. The retrieval results of two models are visualized, including the proposed HGAN model and the HGAN w/o MFR model. HGAN w/o MFR means that we disabled the multi-granularity feature rearrangement (MFR) module of the HGAN model.

In Fig. \ref{figure3}, we show the top 5 retrieval results for three query images. The correct matches are marked with a green ``T”, the wrong matches are marked with a red ``F”, and the whole incorrectly matched sentences are marked in red. The sentences on the left are the retrieval results of the proposed HGAN model, and the sentences on the right are the retrieval results of the HGAN model without MFR. We can observe that the HGAN model can retrieve the correct matching sentence in most cases, and the retrieval accuracy is significantly superior to the latter. In Fig. \ref{figure4}, the top 3 retrieval results for three query sentences are displayed. The correct retrieved images are highlighted in green boxes, while the wrong ones in red boxes. The upper column of images are the retrieval results of the proposed HGAN model, and the lower column of images are the retrieval results of the HGAN model without MFR. 
We can observe that our HGAN model can match the correct image in the top 3 results in Fig. \ref{figure4}. Besides, compared with the HGAN w/o MFR model, the image that matches the query sentence has a higher rank in our HGAN model.

We argue that there are two reasons for this phenomenon. First, the MFR module achieves multi-granularity feature denoising through feature rearrangement, which filters out the noisy parts and retain the dominate parts for the multi-granularity alignment of object-context-fused information.
By mining the multi-granularity semantic relationship of multimodal features, the retrieval performance is improved. Then, the multi-level similarity functions can measure the image and text similarity at different levels in the multi-granularity shared space, which further improves the performance of ITR in terms of hierarchical alignment.

\begin{figure*}[ht]
	\centering
	\includegraphics[width=1\textwidth]{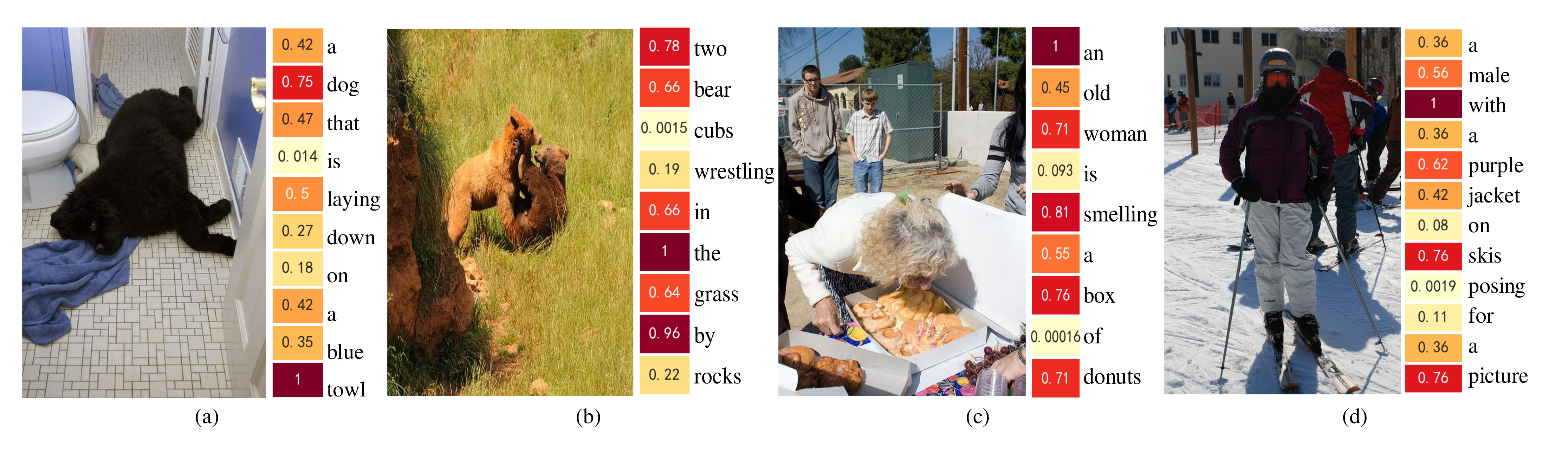}
	\caption{Visualization of word similarity weights. Each subplot shows the similarity (after normalization) between the image and each word in its GT text. (a) and (b) are the samples for which our model retrieves the correct result at top-1. (c) and (d) are samples for which the top-10 items retrieved by our model do not contain correct results, and the incorrect top-1 result given by HGAN is ``A man buying some food at a food stand'' for (c) and ``An adult skier carries a child skier under their arm on the slopes'' for (d), respectively.}
	\label{more-view}
\end{figure*}

\begin{figure*}[ht]
	\centering
	\includegraphics[width=1\textwidth]{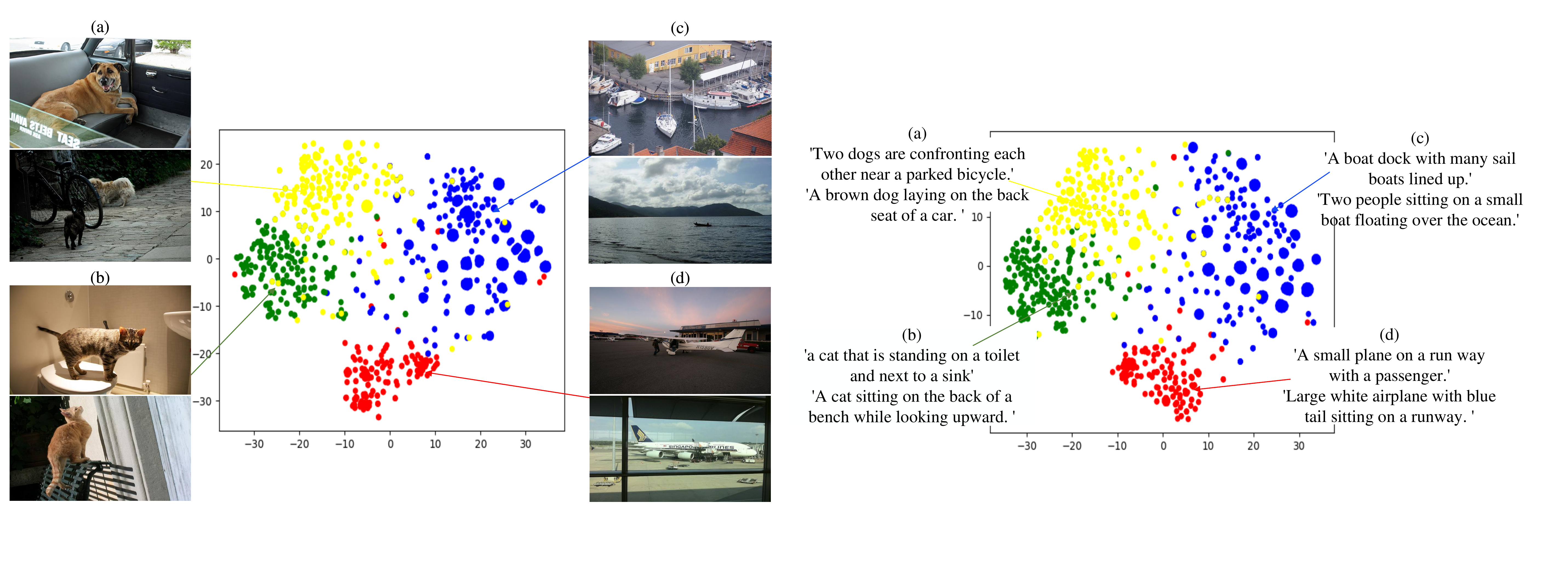}
	\caption{T-SNE visualization of the image feature (left) and the corresponding text feature (right) on a subset of the MS-COCO test dataset. Different colors represent the different classes of samples.}
	\label{tsne}
\end{figure*}

\subsection{Visualization of Features}
In this section, we perform visual analyses of the features modeled by our HGAN model to better demonstrate the effect of our model. \par
First, we display the detailed matching relationships between images and words. Specifically, we calculate the scores of image-word matching for four sample pairs, as shown in Fig. \ref{more-view}, where (a) and (b) are the samples for which our model retrieves the correct result at top-1, (c) and (d) are samples for which the top-10 items retrieved by our model do not contain correct results. The incorrect top-1 result given by HGAN is `A man buying some food at a food stand' for (c) and `An adult skier carries a child skier under their arm on the slopes' for (d), respectively. In each subplot, the selected image is shown on the left, the corresponding GT text is shown on the right, and in the middle is the similarity between the image and each word in its GT text.

All four subplots reveal that our model successfully recognizes the objects with key semantics in the samples, such as ``dog'', ``towl'' in (a) and ``women'', ``donuts'' in (c). Besides, since our model relies heavily on the regions obtained by object detection in constructing the features of images, it has poor understanding of articles, prepositions and verbs. For example, ``the'' in (b) and ``with'' in (d) mistakenly have the maximum similarity with the image.

Observing (c), (d) with their corresponding incorrect top-1 retrieval results, respectively, we can find that they describe a very similar scene, but there are detail errors and focus deviations. For example, ``woman is smelling donuts'' in the picture of (c) is wrongly judged as ``the man is buying food'', and people in the background of the picture in (d) are mistakenly focused.\par
In conclusion, our model has an excellent ability to find objects and nouns containing important semantics in the sample to achieve cross-modal image-text matching. However, the poor understanding of articles, prepositions and verbs can lead to misunderstanding in semantic details, which needs further improvement. 

Then, we conduct a T-SNE visualization experiment by using a subset of the MS-COCO test dataset as shown in Fig. \ref{tsne}. Concretely, we randomly chose samples from the MS-COCO test set with four class labels (aeroplane, boat, cat, and dog) and fed their data into our HGAN model to generate features. The high-dimensional features of the samples are transformed into two-dimensional by T-SNE and displayed as points in Fig. \ref{tsne}, where different colors used to differentiate their classes. For each of these four categories, a set of corresponding image and text samples labeled with (a)$\sim$(d) is displayed.
In each subplot, we can find that the points of each color are aggregated in a single region, indicating that the model has learned the discriminative information belonging to different classes of samples. Comparing the points of same color between the left and right subplots, it is found that they appear in similar areas, indicating that the model achieves cross-modal semantic matching. In addition, the distribution of the yellow and green points is closer because the classes they belong to (cat and dog) are more similar. Moreover, some points appear in the wrong colour area, such as the red points (aeroplanes) that appear in the blue area (boats). We analyze that there are two reasons for this. One is that planes and ships often appear in similar scenes, such as the blue sky and the sea, which leads to the error of model judgment. Another reason is that the selected samples may contain multiple types of objects, so there will be some areas of color mixing in this experiment.

\section{Conclusion}
In this paper, we propose a novel Hierarchical Graph Alignment Network (HGAN) for image-text retrieval, including the following advantages:
 \begin{itemize}
	\item{}We construct feature graphs for the image and text modalities respectively to capture more comprehensive multi-modal features, and establish a multi-granularity shared space with the designed Multi-granularity Feature Aggregation and Rearrangement (MFAR) module to achieve multi-granularity feature filtering and fusion.
	\item{}We establish hierarchical alignment across modalities for features of varying granularity using three-level similarity functions, which deeply explore the feature similarity in the multi-granularity shared space.
	\item{}Extensive experiments on the MS-COCO and Flickr30K datasets show that the proposed HGAN method outperforms the state-of-the-art models for the ITR task. 
\end{itemize}
In the future, for a more comprehensive feasibility analysis of the model, we are ready to extend our model to more tasks, such as image captioning \cite{Yu2022DualAO} and visual question answering \cite{Jiang2020InDO}, as well as more modalities, like video-text field \cite{Wang2021LearningCG}.


\section*{Acknowledgments}
This work was supported by the National Natural Science Foundation of China under Grant (Nos. 62071354, 62201424 and 61906156), China Postdoctoral Science Foundation Grant (No. 2017M620438), the Natural Science Foundation of Shaanxi Province (Nos. 2019ZDLGY03-03) and also supported by the ISN State Key Laboratory and High-Performance Computing Platform of Xidian University.

\bibliographystyle{IEEEtran}
\bibliography{HGAN}


%
%
%
%

\vfill

\end{document}